\documentclass{article}
\usepackage{microtype}
\usepackage{graphicx}
\usepackage{booktabs}
\usepackage{caption}
\usepackage{xparse}
\usepackage{bbding}
\usepackage{colortbl}
\usepackage{bbm}
\usepackage[utf8]{inputenc}
\usepackage[table,xcdraw]{xcolor}
\usepackage{colortbl}
\usepackage{makecell}
\usepackage{booktabs}
\usepackage{amsmath}
\usepackage{hyperref}
\usepackage{enumitem}
\usepackage{adjustbox}
\definecolor{champagne}{rgb}{0.74, 0.83, 0.9}
\definecolor{champagne}{rgb}{0.97, 0.91, 0.81}
\definecolor{gray!20}{gray}{0.8}
\usepackage{hyperref}

\usepackage[accepted]{icml2024}
\usepackage{amsmath}
\usepackage{amssymb}
\usepackage{mathtools}
\usepackage{amsthm}
\usepackage{algorithmic}
\usepackage{algorithm}
\usepackage{subcaption}
\usepackage[capitalize,noabbrev]{cleveref}

\Crefname{section}{Sec.}{Sec.}
\Crefname{subsection}{Sec.}{Sec.}
\Crefname{algorithm}{Alg.}{Algs.}
\definecolor{darksalmon}{rgb}{0.91, 0.59, 0.48}
\definecolor{emerald}{rgb}{0.31, 0.78, 0.47}
\definecolor{green(pigment)}{rgb}{0.0, 0.65, 0.31}
\definecolor{amaranth}{rgb}{0.9, 0.17, 0.31}
\definecolor{iris}{rgb}{0.35, 0.31, 0.81}
\definecolor{uu}{rgb}{0.95, 0.51, 0.51}
\definecolor{spirodiscoball}{rgb}{0.06, 0.75, 0.99}

\hypersetup{
  breaklinks,
  colorlinks=iris,
  linkcolor=amaranth,
  citecolor=iris,
  urlcolor=uu,
}

\DeclareMathOperator*{\argmax}{arg\,max}
\DeclareMathOperator{\R}{\mathbb{R}}

\DeclareMathOperator{\E}{\mathbb{E}}

\DeclareMathOperator{\pre}{pre}

\newcommand{\savefont}{\xdef\oldfontsize{\f@size}\xdef\oldblskip{\f@baselineskip}}



\theoremstyle{plain}

\theoremstyle{definition}

\theoremstyle{remark}

\NewDocumentCommand\gptswarm{}{{\raisebox{-1.5mm}{\includegraphics[scale=0.075]{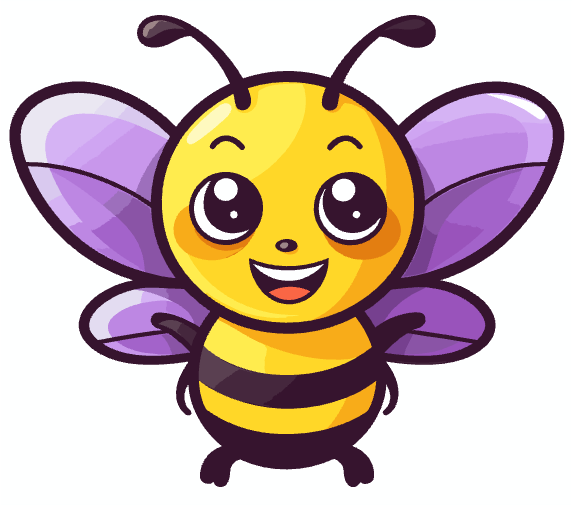}}}}

\usepackage[textsize=tiny]{todonotes}
\icmltitlerunning{GPTSwarm: Language Agents as Optimizable Graphs}

\begin{document}
\twocolumn[
\icmltitle{GPTSwarm: Language Agents as Optimizable Graphs}
\icmlsetsymbol{equal}{*}

\begin{icmlauthorlist}
\icmlauthor{Mingchen Zhuge}{equal,kaust}
\icmlauthor{Wenyi Wang}{equal,kaust}
\icmlauthor{Louis Kirsch}{idsia}
\icmlauthor{Francesco Faccio}{kaust,idsia}
\icmlauthor{Dmitrii Khizbullin}{kaust}
\icmlauthor{Jürgen Schmidhuber}{kaust,idsia}
\end{icmlauthorlist}




\icmlaffiliation{kaust}{AI Initiative, King Abdullah University of Science and Technology (KAUST), Thuwal, Saudi Arabia}
\icmlaffiliation{idsia}{The Swiss AI Lab IDSIA, USI, SUPSI, Lugano, Switzerland}

\icmlcorrespondingauthor{Mingchen Zhuge}{mingchen.zhuge@kaust.edu.sa}
\icmlcorrespondingauthor{Dmitrii Khizbullin}{dmitrii.khizbullin@kaust.edu.sa}

\begin{center}
    \gptswarm \; \url{https://gptswarm.org}
\end{center}

\icmlkeywords{Machine Learning, ICML}

\vskip 0.3in
]

\begin{figure*}[!htbp]
\centering
\includegraphics[width=0.98\textwidth]{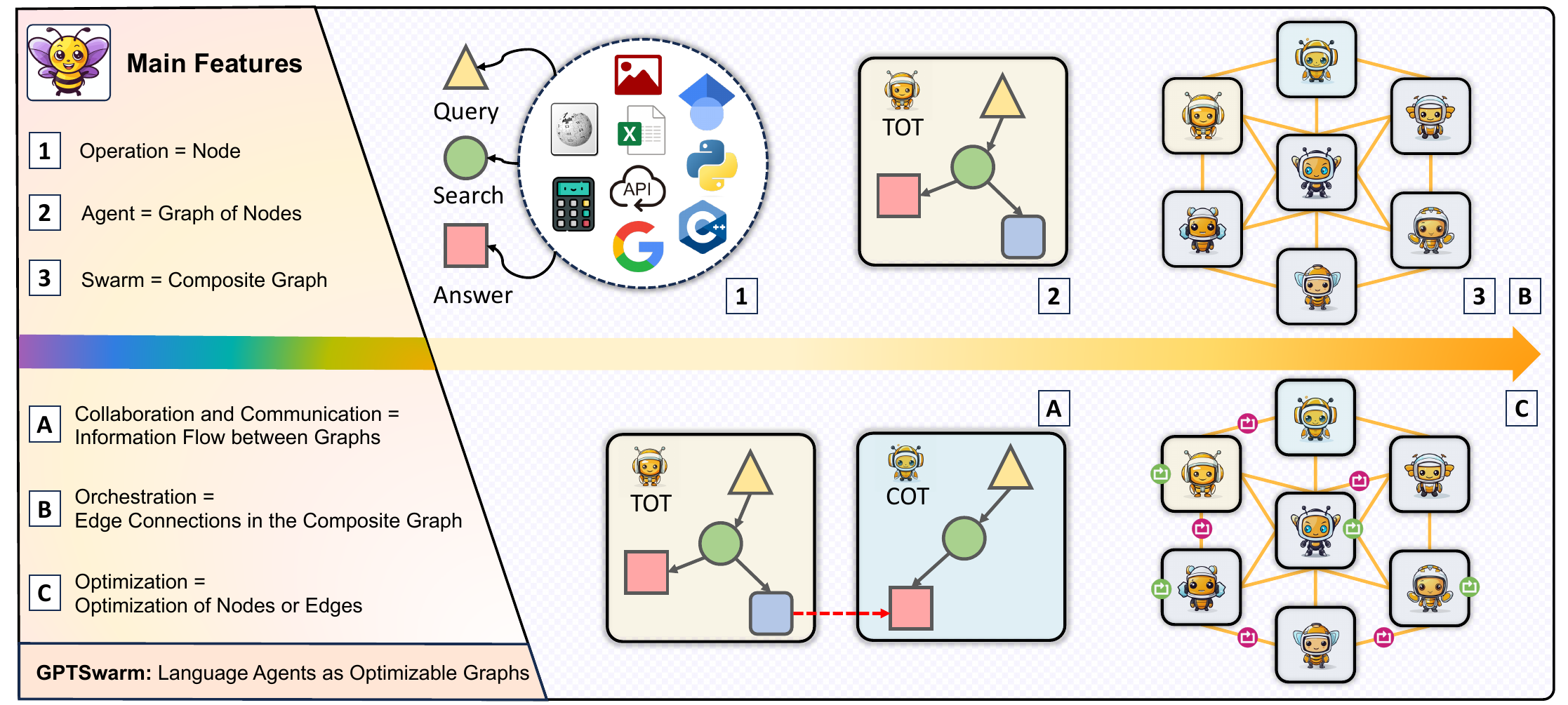}
\vspace{-7pt}
\caption{
\textbf{GPTSwarm is a framework that represents agents as graphs.} 
In this framework, each node represents an operation (e.g., LLM inference or tool use).
An agent is a graph composed of these nodes.
An edge between two agent graphs characterizes a communication channel; each agent collaborates with others through different channels.
When connected, multiple agents form a composite graph with a certain orchestration topology.
This graph representation lends itself to optimization of nodes and edges via prompting and evolutionary or reinforcement learning techniques.
}
\label{fig:teaser}
\end{figure*}

\printAffiliationsAndNotice{\icmlEqualContribution}

\begin{abstract}

Various human-designed prompt engineering techniques have been proposed to improve problem solvers based on Large Language Models (LLMs), yielding many disparate code bases.
We unify these approaches by describing LLM-based agents as computational graphs.
The nodes implement functions to process multimodal data or query LLMs, and the edges describe the information flow between operations. 
Graphs can be recursively combined into larger composite graphs representing hierarchies of inter-agent collaboration
(where edges connect operations of different agents).
Our novel automatic graph optimizers (1) refine node-level LLM prompts (node optimization) and (2) improve agent orchestration by changing graph connectivity (edge optimization).
Experiments demonstrate that our framework can be used to efficiently develop, integrate, and automatically improve various LLM agents.
The code can be found \href{https://github.com/metauto-ai/gptswarm}{here}.
\end{abstract}

\section{Introduction}

Interest in LLM-powered autonomous problem solvers or agents and their varied applications is continually rising~\citep{wang2023survey,xi2023rise}.
However, much work remains to be done to effectively incorporate these agents into a cohesive society and improve their structure automatically.

Early approaches zero-shot-prompted LLMs or prompted them with few-shot examples~\citep{kojima2022large,brown2020language}. 
Recent methods prompt LLMs in a structured way, such as chain of thought (COT)~\citep{wei2022chain}, ReAct~\citep{yao2022react}, tree of thought (TOT)~\citep{yao2023tree}, Reflexion~\citep{shinn2023reflexion}, and Graph of Thought (GOT)~\citep{besta2023graph}, to improve text-based reasoning.
Single agent applications such as AutoGPT~\citep{autogpt_github}, BabyAGI~\citep{nakajima2023babyagi}, LangChain~\citep{Chase_LangChain_2022}, and Llama-index~\citep{Liu_LlamaIndex_2022} utilize LLMs for various functionalities, including tool usage, function calling, and embodied actions.
In multi-agent frameworks~\citep{zeng2022socratic,zhuge2023mindstorms} several LLMs take on different roles~\citep{li2023camel,park2023generative,qian2023communicative,wu2023autogen} 
to communicate in natural language and collectively solve a given task.
This approach often outperforms single agents, exploiting the specialization~\citep{hong2023metagpt} of various LLM agents. 
Unfortunately, it also leads to increasingly different and disparate code bases that require a lot of human engineering to define prompting schemes and the workflow of agents.

In a ``society of mind" (SOM)~\citep{minsky1988society,zhuge2023mindstorms}, higher-level intelligence emerges from the combination of simpler and modular cognitive components. Inspired by SOMs, we describe language agent systems through graph representations.
Language agents querying LLMs and utilizing external tools are modeled as computational graphs where each node is dedicated to a specific function, while the edges define a topology of how inputs are processed across nodes, mirroring the prompting schemes in prior studies.
A swarm is defined as a composite graph, where each subgraph represents a collaborative agent.
This creates a deeper hierarchy of intelligence. Agent graphs combine basic LLM operations~\citep{kennedy2006swarm,nepusz2013hierarchical}, and swarm graphs contain subgraphs representing agents.
Approaches such as COT~\citep{wei2022chain}, TOT~\citep{yao2023tree}, and Self-Consistency~\citep{wang2022self} can be represented by our graphs.

Our graph representation lends itself to optimization via prompting and evolutionary or reinforcement-learning techniques, so that agents can improve their communication (or orchestration) patterns. The graph connectivity (adjacency matrices) between agents can self-improve online as a task is being solved or its solution is transferred to another task.

As a proof-of-concept, we demonstrate how suboptimal agent organization can be overcome and how existing prompting techniques, such as Tree of Thought and Reflexion, can be automatically recombined by optimizing edges in a composite graph.
Apart from edge optimization, our framework allows each node in the graph to self-improve by adapting its prompts based on previous input and task feedback.

\textbf{Our contributions can be summarized as follows:}

\noindent \textbf{(1)} We unify language agent systems by describing them as optimizable computational graphs.
\vspace{-3pt}

\noindent \textbf{(2)} We introduce an open-source framework that allows for constructing arbitrary agent systems by recombining fundamental operations.
We describe these engineering-level contributions in \Cref{apendix:gptswarm_engineering}.
\vspace{-3pt}

\noindent \textbf{(3)}  We develop optimization methods for nodes and edges, enabling automatic improvements of agent prompts and inter-agent orchestration.
\vspace{-3pt}

\noindent \textbf{(4)} We validate our framework on various benchmarks including MMLU, Mini CrossWords, HumanEval, and GAIA,  with an emphasis on the benefits of automatic graph optimization.

\section{GPTSwarm}

\subsection{Language Agents as Graphs}

Taking inspiration from the society of mind (SOM)~\citep{minsky1988society, zhuge2023mindstorms}, we propose to organize intelligence within a modular and hierarchical framework.
This framework consists of nodes, graphs, and composite graphs, with each component playing a specific role.
A node represents a fundamental operation that includes, but is not limited to, LLM inference, tool use, function calls, and various embodied actions.
An agent, conceptualized as a graph, consists of multiple nodes that form a coherent functional entity.
A swarm, or composite graph, represents a complex system of agents where the collective capabilities of this system may exceed those of individual agents. 
Finally, the edges within an agent define its execution topology, while the edges between agents establish collaboration and communication among them.

\subsection{Graph Definition}

\paragraph{Single language agent as a graph.} We model a language agent as a directed computational graph $G$, defined by a tuple $(N, E, F, o)$, where $N$ is a set of computational nodes, $E \subset N \times N$ is a set of directed edges, $F = \{f_n\}_{n\in N}$ is a set of computational routines and $o \in N$ is an output node.
The set of predecessors of node $n$ is denoted by $\pre(n)$.
In this paper, we focus on directed acyclic graphs (DAGs). 
Given an input $x$, a graph $G$ iteratively executes its nodes according to their topological order. 
Each node $n \in N$ receives as input $x$ and the output $z_n$ from its predecessor nodes.
In this work, inputs and outputs are strings in natural language, but may take on other data types more generally.
Node $n$ applies the computational routine $f_n (z_n, x)$ and sends the output to its successor nodes.
The graph output, denoted $\hat{y} = G(x)$, is the output $f_o (z_o, x)$ from the output node $o$.
Note that in a DAG, some nodes will not have predecessors.
For such nodes, the context $z$ will be empty.
This graph execution procedure is summarized in Algorithm~\ref{alg:graph_exe}.

\vspace{-5pt}
\begin{algorithm}[h]
\begin{algorithmic}
\REQUIRE Computational graph $G = (N, E, F, o)$, input $x$, empty context $z$ for each node without predecessors.
\FOR{$n$ in $\text{TopologicalSort}(N)$}
\STATE $z_n \leftarrow \{f_v(z_v, x): v \in \pre(n)\}$ 
\ENDFOR
\ENSURE $f_o(z_o, x)$
\end{algorithmic}
\caption{Graph Execution}\label{alg:graph_exe}
\end{algorithm}

In the context of language agents, for example, the input $x$ may correspond to a question in natural language.
Each node processes the input $x$ and context information $z$ from its predecessor nodes by applying a computational routine $f$.
Examples of routines include LLM queries with input data from other agents, instructions to generate prompts for web searches that gather task-related information, or tool usage.
Although our formalization specifies that the input $x$ is given to each node, in practice, many routines might be designed to ignore the input and operate solely in the context provided by the predecessor nodes.
Finally, the output provided by the output node corresponds to the answer to the input question or, more generally, to the solution of the input task. 

\paragraph{Swarm of language agents as a composite graph.} Given a set of $K$ language agents, each represented by a computational graph $\{G_k=(N_k, E_k, F_k, o_k)\}_{k=1}^K$, one can compose these agents to achieve high performance in specific tasks.
Let $N'=\cup_k N_k$ represent the union of the nodes of the agents, $E' = \cup_k E_k$ be the union of the edges of the agents, $F'=\cup_k F_k$ be the union of the computational routines of the agents, and $o'\in \cup_k\{o_k\}$ be the output node for the composite graph.
Consider a selection of edges $\mathcal{E} \subset \cup_{i\neq j} N_i \times N_j$ that describe a set of connections between nodes from different agents.
We define the composite graph representing the swarm of agents as $G_{\mathcal{E}}=(N', E_{\mathcal{E}}, F', o')$, where $E_{\mathcal{E}} = E' \cup \mathcal{E}$ is the union of the edges of the agents and the new edges connecting them.
Composite graphs are restricted to DAGs.
The composite graph $G_{\mathcal{E}}$ can be executed as described in Algorithm~\ref{alg:graph_exe}.
In a swarm of language agents, the newly specified edges represent communication channels between agents.
In the following sections, we explore how to optimize such a computational graph.

\subsection{Edge Optimization}
Given a task $\tau$ and its associated utility function $u_\tau$ that maps the candidate graphs to real numbers,
we formulate an optimization problem about the choice of additional edges. 
The goal is to identify the edges that connect various language agents in a swarm, maximizing the utility. This process involves determining the most effective patterns of communication and information exchange among agents for the task at hand.
We consider a set of potential edges $\{e_i\}_{i=1}^d = \mathcal{E}$, which leads to $2^d$ possible edge configurations, symbolized as $\mathcal{E} \in \{0,1\}^d$. We further restrict the search space to only consider composite graphs that are DAGs.
Formally, optimization of the composite graph of language agents is achieved by solving the problem $\max_{\mathcal{E}} u_{\tau}(G_{\mathcal{E}})$.

\subsubsection{Problem Reformulation}
DAG optimization through pruning of nodes and edges was already present in the first work on ``deep learning" with deep feedforward networks~\citep{ivakhnenko1965cybernetic, ivakhnenko1968group}.
Due to the combinatorial complexity induced by DAGs, recent studies have increasingly focused on the continuous optimization approach~\citep{vowels2022d}. 
This is particularly relevant in scenarios where most node executions require one or more queries to LLMs for moderate-scale applications. 
Moreover, the utility function is typically non-differentiable due to the tokenization of LLMs, and this remains true even when a differentiable DAG sampling technique is employed.
Therefore, we reformulate our edge optimization as a continuous optimization problem.
Instead of optimizing in a discrete space, our approach is to optimize over a continuum of probabilistic distributions, each representing a distribution over the feasible DAGs.
Formally, rather than solving the maximum utility function $\argmax_{\mathcal{E}} u_\tau(G_{\mathcal{E}})$, we propose solving
\begin{align}
    \argmax_{\theta \in \Theta} \E_{G' \sim D_\theta}[u_\tau(G')],
    \label{eq:objective}
\end{align}
where $D_\theta$ is a parameterized distribution and $\Theta$ represents a feasible set of real-valued parameters.

\subsubsection{Solution Parameterization}
A straightforward way to define a parameterized probabilistic distribution over DAGs with fixed nodes $N$ and required edges $E$ is to assign a real-valued parameter $\theta_i \in \R$ to each potential edge $e_i$. Let $\theta = [\theta_1;\, \theta_2;\, \dots;\, \theta_d] \in [0, 1]^d$. The probability of $G' = G_{\mathcal{E}}$ for $G' \sim D_\theta$ is 
\begin{align*}
    \prod_{i=1}^d\begin{cases}
    \theta_i \quad &\text{if}\,
    (N, E\cup(\{e_j\}_{j=1}^{i-1} \cap \mathcal{E}) \cup \{e_i\}) \,\text{is a DAG,}\\
    0 &\text{otherwise.}
    \end{cases}
\end{align*}
A sampling method that realizes this distribution is first to initialize a graph $G' \leftarrow (N, E)$. Then, iteratively sample whether to include edge $e_i$ in $G'$ for all $i$'s. If including $e_i$ causes a cycle in current $G'$, then the edge would not be included. Otherwise, add the edge to $G'$ with probability $\theta_i$.

\subsubsection{Optimization Algorithm}
To optimize the objective function (\Cref{eq:objective}), we apply the REINFORCE algorithm~\citep{williams1992simple} by applying a gradient ascent variant (e.g., Adam \citep{kingma2014adam}) with an unbiased gradient estimation:
\small
\begin{align}
\nabla_\theta \E_{G_{\mathcal{E}} \sim D_\theta\ }[u_\tau(G_{\mathcal{E}})] \approx \frac{1}{M} \sum_{i=1}^M\hat{u}_\tau(G_i) \nabla_\theta \log(p_\theta(G_i)),
\label{eq:grad}
\end{align}
\normalsize
where $G_1, G_2, \dots, G_N \sim D_\theta$ are mutually independent and $\hat u_\tau(G_i)$ is an independent unbiased estimate of $u_\tau(G_i)$ for all $i$ and some $M \in \mathbb{N}$.
Algorithm \ref{alg:reinforce} describes the optimization algorithm with vanilla gradient ascent.

\begin{algorithm}[h]
\begin{algorithmic}
\REQUIRE A parameterized probabilistic distribution over computation graphs $D_\theta$, an unbiased utility estimator $\hat u_\tau(\cdot)$, and a learning rate $\alpha$.
\STATE Initialize $\theta \in \R^d$.
\WHILE{terminate condition not met}
\STATE Sample $G_i \sim D_\theta$ for $i = 1, 2, \dots, M$.
\STATE Update $\theta \leftarrow \theta + \frac{\alpha}{M} \sum_{i=1}^M\hat{u}_\tau(G_i) \nabla_\theta \log(p_\theta(G_i))$.
\ENDWHILE
\end{algorithmic}
\caption{Edge Optimization with REINFORCE}\label{alg:reinforce}
\end{algorithm}

\subsection{Node Optimization}

\label{sec:node_opt}
In our framework, each node implements a fundamental operation, such as querying an LLM, using a tool, calling an API, etc. 
In a language agent, most of these operations involve prompting an LLM once or several times.
Optimizing the prompts of these nodes is crucial for improving the system's overall performance.

Unlike many other works on prompt optimization, which optimize a single global prompt~\citep[e.g.,][]{yang2023large,pryzant2023automatic,deng2022rlprompt}, our node optimization problem naturally involves several operations where each of them consists of a node-level prompt.
In our case, the optimization problem is more complex due to prompts affecting how other prompts operate on connected nodes.
At the same time, our graph representation leads to a separation of concerns where each node has a specific purpose with its own associated prompt.
Due to this separation of concerns, we hypothesize that, for every optimization step, it is sufficient to update each node-level prompt individually, assuming that all other prompts are fixed.

Consider a parameterized computational graph $G^P = (N, E, F^P, o)$, where $F^P = \{f_n^{p_n}\}$ are computational routines, each parameterized by a prompt $p_n$ to be optimized for all $n \in N$.
To enable effective node optimization, we also require a natural language description of the intended function for each routine $f_n^{p_n} \in F$ denoted by $d_n$.
For example, a suitable description for a node designed to write Python programs would be ``a Python code generator".
Here, existing prompt optimization methods, such as OPRO~\citep{yang2023large}, can be described as a function $I$ that iteratively maps a prompt, a function description, and a set of node input-output pairs (which may include annotations such as a quality measure for each pair) to an improved prompt.
For example, $I$ could take a prompt such as ``generate Python code", a description ``a Python code generator", and an input-output pair ``Input: evaluate two divided by one as an integer. Output: 2 / 1", where the output yields $1.0$ as the result of execution. A prompt optimization method would return an improved prompt ``generate Python code and pay attention to data types".

Formally, our method begins by initializing an empty history set, denoted $h_n$, one for each node $n \in N$.
The process then proceeds iteratively: first, the graph $G^P(x)$ is executed using a randomly sampled input $x$ following Algorithm \ref{alg:graph_exe}.
Subsequently, for each node, a tuple consisting of the input to the node $(z_n, x)$, where $z_n$ is the context vector that includes the outputs of the predecessor nodes, and the node's own output $f_n^{p_n}(z_n, x)$, is added to the node's history $h_n$.
The final step involves updating the node prompts.
This is done by applying $I$  to the node's updated history, its current prompt, and its function description, resulting in an improved prompt $I(h_n, p_n, d_n)$.
This iterative process, described in Algorithm \ref{alg:node_opt}, continuously improves the operations of the nodes in the entire graph.

\begin{algorithm}[h]
\begin{algorithmic}
\REQUIRE A parameterized graph $G^P = (N, E, F^P, o)$, natural language function descriptions $D = \{d_n\}_{n \in N}$, and a distribution of inputs $D_X$.
\STATE Initialize $p_n$ for all $n \in N$.
\STATE Initialize $h_n \leftarrow \emptyset$ for all $n \in N$.
\WHILE{terminate condition not met}
\STATE Sample input $x \sim D_X$.
\STATE $y \leftarrow G^P(x)$ following Algorithm \ref{alg:graph_exe}.
\STATE $h_n \leftarrow h_n \cup \{((z_n, x), f_n^{p_n}(z_n, x))\}$ for all $n \in N$.
\STATE $p_n \leftarrow I(h_n, p_n, d_n)$, for all ${n\in N}$.
\ENDWHILE
\end{algorithmic}
\caption{Node Optimization}\label{alg:node_opt}
\end{algorithm}
\vspace{-5pt}

\subsection{General Applicability}

Frameworks such as AutoGPT~\citep{autogpt_github} and LangChain~\citep{Chase_LangChain_2022} have set a standard for flexibility and reusability in various language-based tasks.
Our framework, GPTSwarm, introduces a graph-based design of agents and swarms.
This design further simplifies the reuse of modular components (nodes \& agents) and the integration of such modules.
For instance, GPTSwarm supports 41 types of file analysis, web search (e.g., Google Search), and index-based memory.
By offering a wide range of modules, our framework makes it easier to implement various language agent systems.
See Section \ref{sec:gaia} for further details.

\section{Experiments}

\subsection{MMLU}
\subsubsection{Adversarial setting}
\label{sec:adversarial}
\noindent \textbf{Motivation:}
In our first experiments, we demonstrate that edge optimization effectively filters adversarial agents from a swarm, mirroring a scenario in multi-agent systems where some agents are detrimental rather than beneficial.
Ideally, optimization would automatically eliminate harmful agents.
We conducted this experiment using the 4-choice MMLU general knowledge question answering dataset, as detailed by \citet{hendryckstest2021, hendrycks2021ethics}.
Our setup involves initializing a swarm consisting of $k$ Input-Output (IO) agents and $k$ adversarial agents, following the terminology by \citet{besta2023graph}.
The IO agents query an LLM and relay the LLM's responses directly.
In contrast, adversarial agents are deliberately programmed to manipulate the LLM to provide incorrect answers.
The collective decision on the final answer is made through majority voting, bypassing any additional LLM query that could introduce corrective intelligence against adversarial influence.
We benchmark the performance of a single IO agent as our baseline. An effective optimization is expected to elevate the swarm's performance on the MMLU dataset to match this baseline level.

\begin{figure}[t!]
\centering
\vspace{-9pt}
\includegraphics[width=0.45\textwidth]{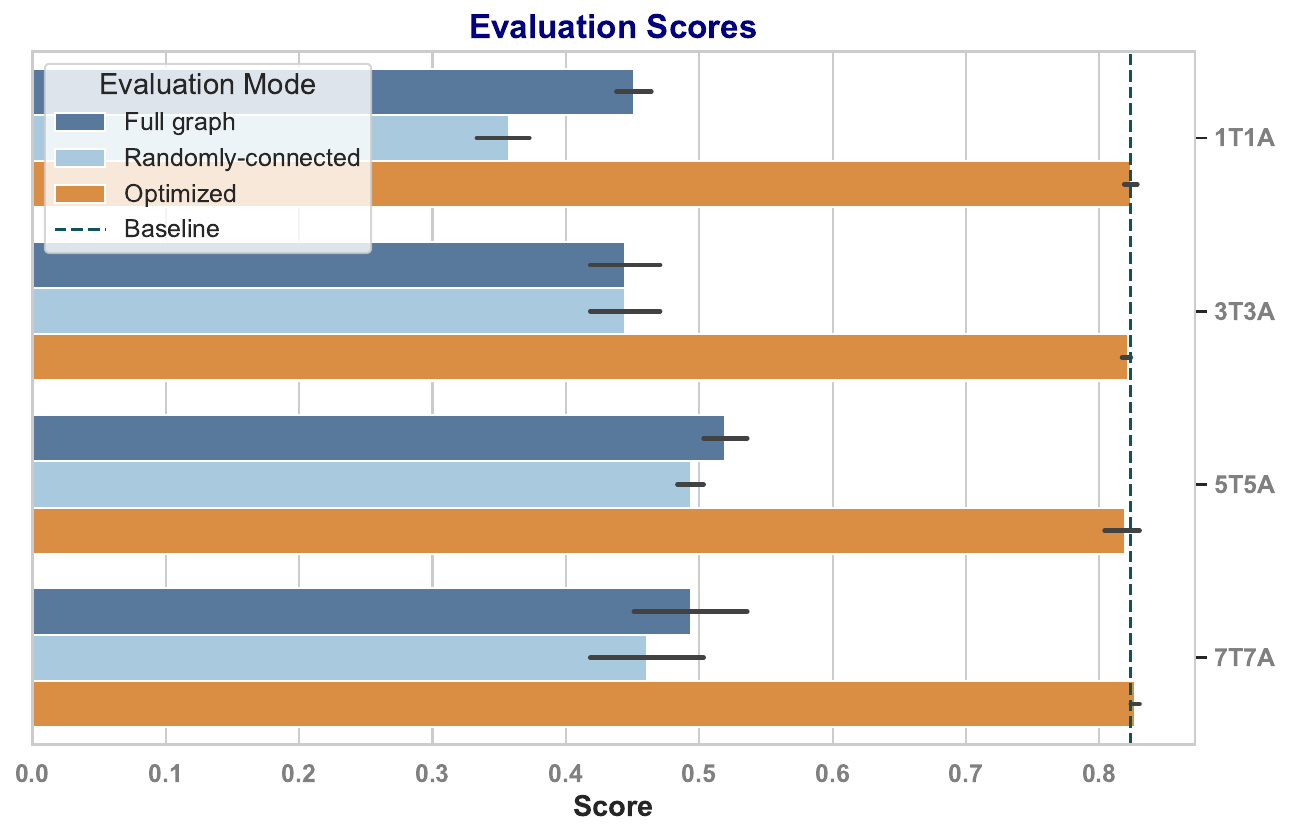}
\vspace{-15pt}
\caption{
\textbf{Score recovery through edge optimization.}
``T" denotes truthful and ``A" adversarial agents, e.g., a 3T3A swarm has 3 of each.
Ablation studies include a ``full graph" and random graphs sampled according to distribution $D_{\mathbf{0.5}}$.
The dashed line corresponds to the direct answer baseline.
}
\label{fig:adv_mmlu}
\end{figure}

\noindent \textbf{Analysis:}
In Figure \ref{fig:adv_mmlu}, we present the comparative performance scores of different swarm configurations: the baseline, the graph formed by sequentially including edges that do not create loops (denoted as the `full graph'), a randomly connected swarm sampled from the initial distribution $D_{\theta}$ with $\theta=0.5$, and the optimized swarm.
These scores are derived from evaluating the initial 10\% of the MMLU validation set.
The edge optimization process applies REINFORCE (\Cref{alg:reinforce}) for 200 iterations.
Each iteration assesses four graph samples, each on a specific problem sourced from the MMLU dev set.
In all experiments, we used GPT-4-Turbo with the token sampling temperature $0.2$.
Figure \ref{fig:adv_training} demonstrates how the optimized swarm score aligns asymptotically with that of the baseline.
Table \ref{tab:adv_stats} compiles the key statistics and findings of these experiments. 
The findings indicate that our approach successfully safeguards a swarm against harmful adversaries.

\subsubsection{Collaborative setting}
\label{sec:collaborative}
In the adversarial setting above, we observe no score improvement over the single agent baseline.
We attribute this to the fact that all IO agents are prompted identically.
To elicit score improvement, we run a set of 7 different IO agents instructed to behave according to various roles.
See \Cref{apendix:collab_case} for the list of roles.
The optimized swarm improves over the baseline from \Cref{sec:adversarial} by 2.1\% $\pm$ 1.1\% averaged over 5 training seeds.

\subsection{Mini Crosswords}
\label{sec:crosswords}
\noindent \textbf{Motivation:}
This section investigates to what extent edge optimization can improve the performance of standard agents from the literature.
We conduct our evaluation on the Mini Crosswords dataset\footnote{\url{https://www.goobix.com/crosswords}}. A subset of 20 problems is used to optimize and evaluate our methods, in agreement with previous studies~\citep{yao2023tree, sel2023algorithm}.
The choice of Mini Crosswords for this analysis is strategic, as it highlights how the algorithmic structure of the solvers, such as the tree search employed by TOT, significantly influences their performance~\citep{yao2023tree}.
Our hypothesis is that edge connections can meaningfully determine the algorithmic structure.
Through edge optimization, we anticipate the automatic discovery and implementation of high-performance algorithms.

\begin{figure}[t]
\centering
\vspace{4pt}
\includegraphics[width=0.45\textwidth]{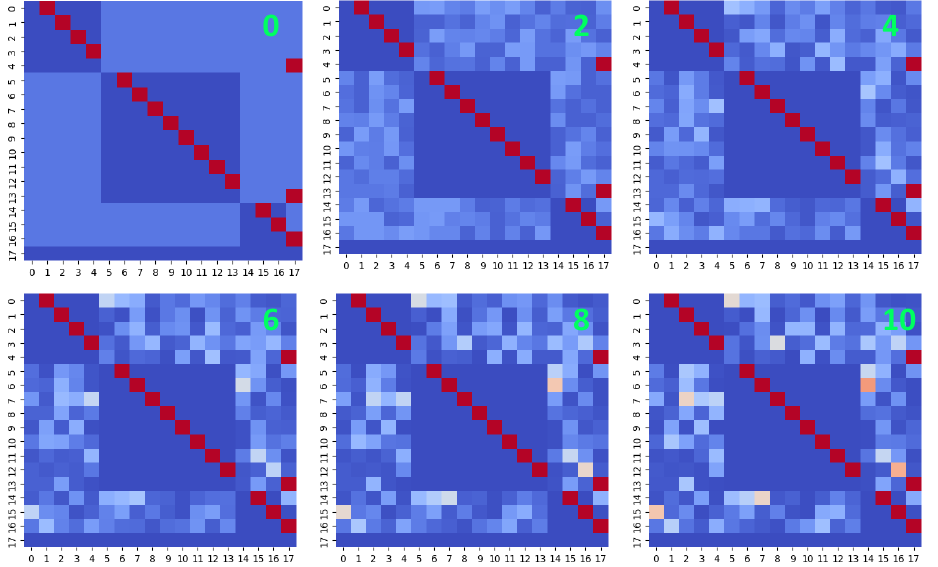}
\vspace{-10pt}
\caption{
\textbf{Visualizing the evolution of the probability distribution during optimization in adjacency-like matrices.}
In this figure, we show the probability parameters (one corresponds to an edge) in an adjacency-like matrix for iterations 0, 2, 4, 6, 8, and 10 of optimizing the objective for the Mini Crosswords task.
We observe that the parameters first change chaotically.
However, after iteration 6, the parameters change almost monotonically. 
}
\label{fig:cw_heatmap}
\vspace{-15pt}
\end{figure}

\noindent \textbf{Analysis:}
In our experiments, we explore the performance of swarms of three distinct agents.
The first agent, which implements the TOT approach, iteratively branches over candidate solutions provided by an LLM, processing one word at each step.
The second agent is based on the Reflexion method \citep{shinn2023reflexion}.
This agent first proposes a solution through a greedy approach and then creates an alternative solution informed by feedback from a critic, which is based on an LLM analysis of the initial solution.
The third agent we examine is a Chain of Thought (COT) agent consisting of three nodes.
Each node within the COT performs an internal brute-force search to select the optimal subset of candidates generated by the LLM for the current state, scored by the LLM.
The agent or swarm then returns all the solutions generated by their output node.

For the utility function, we choose the best of all the graph-returned solutions according to the number of words correctly filled (i.e., best state word accuracy) as done by \citet{yao2023tree}.
During the evaluation, we average over 20 graph samples from the graph distribution, each evaluated on a unique question randomly sampled from the dataset.

We optimize our composite graph of agents using the REINFORCE (Alg. \ref{alg:reinforce}), setting the initial edge probability to $\theta = 10\%$ and the learning rate to $\alpha = 0.4$.
For each iteration, the gradient is estimated according to equation (\ref{eq:grad}) by sampling $M = 20$ graphs, each evaluated on a crossword problem.
For cost-effectiveness, we optimize and evaluate graphs with
the GPT-3.5-Turbo language model, where the temperature is set to zero.
Figure \ref{fig:cw_heatmap} visualizes the evolution of probability parameters in the form of adjacency-like matrices over ten iterations.
We observe that the parameters first change chaotically.
However, after iteration 6, the parameters change almost monotonically. 

We follow Alg. \ref{alg:reinforce} to optimize the objective in Equation \ref{eq:objective}, achieving an average accuracy of $0.575 (\pm 0.0275)$ after ten iterations (we report the average over $3$ runs and the standard error).
This surpasses the initial distribution's score of $0.465 (\pm 0.0509)$.
Furthermore, we evaluate the best-of-three performance by aggregating the top results from each problem across the three agents, which yields an accuracy of $0.320 (\pm 0.0415)$.

Note that denser graphs are likely to require more computational resources.
To verify that the improvements of our method are not solely due to an increase in the number of edges and therefore a larger computational budget, we compare it with a distribution with all parameters set to $\theta = 12.5\%$.
This value reflects the average number of edges in the learned distribution, determined by sampling 1000 graphs from each run's resulting distribution.
The expected number of edges for both the learned distribution and the $0.125$ parameter-valued distribution are approximately $32.76 (\pm 1.93)$ and $32.80 (\pm 0.11)$, respectively.
Despite the similarity in the edge count, the $0.125$ parameter-valued distribution achieves an accuracy of $0.510 (\pm 0.0552)$, allowing us to attribute the improvements to factors beyond the mere edge density.

\begin{figure}[t!]
\centering
\vspace{-5pt}
\includegraphics[width=0.5\textwidth]{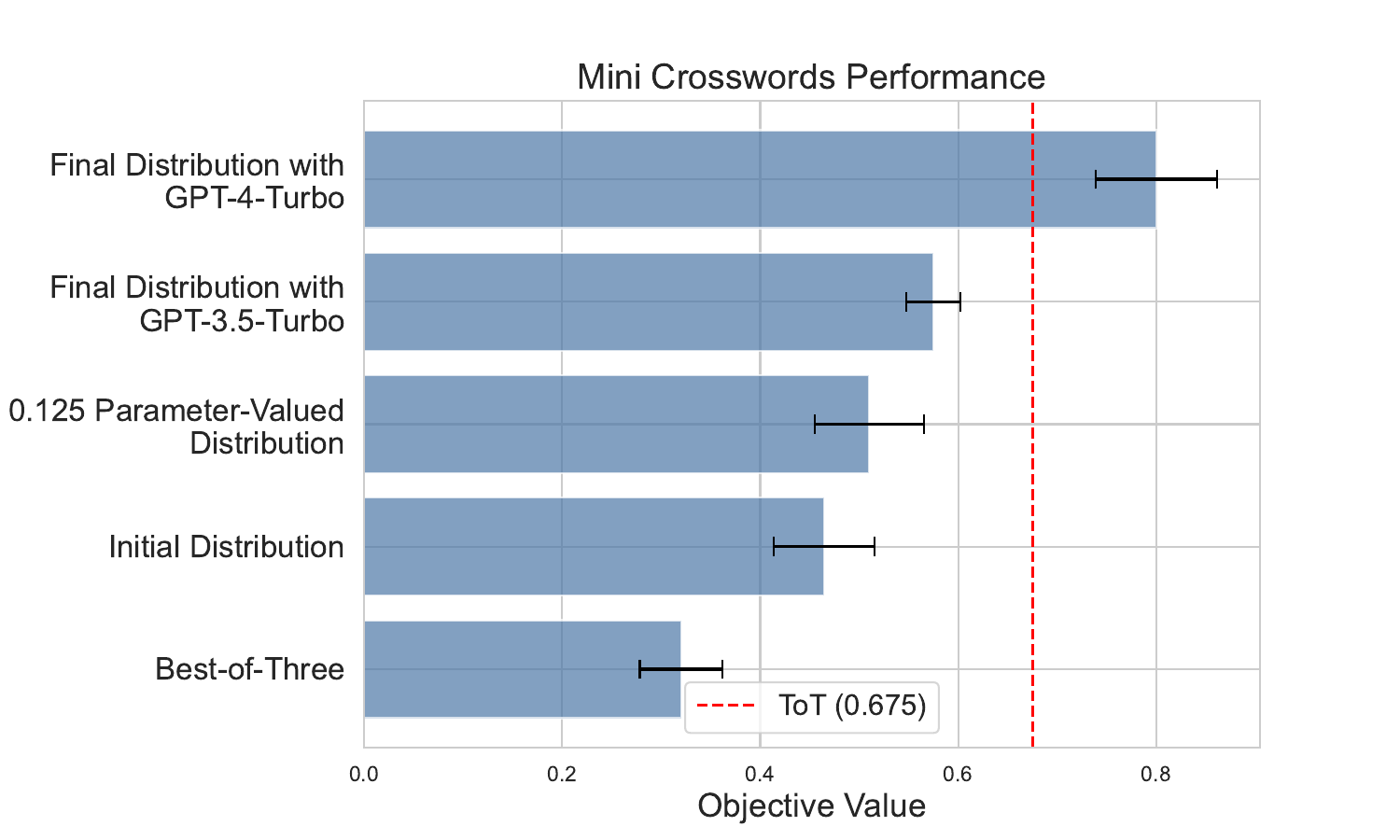}
\caption{
\textbf{Edge optimization on the Mini Crosswords dataset improves over standard methods}
The baseline methods are evaluated with GPT-3.5-Turbo.
The optimized final distribution outperforms several baselines.
When evaluating the already optimized edge distribution with GPT-4-Turbo, we achieve better results compared to the previous state-of-the-art method (Tree of Thought evaluated with GPT-4).
}
\label{fig:crosswords}
\vspace{-10pt}
\end{figure}

Furthermore, we evaluate one of our final optimized distributions (randomly selected) with the GPT-4-Turbo language model\footnote{We are limiting our evaluation to a single graph distribution due to the high cost associated with API calls for this type of evaluation.} (Figure \ref{fig:crosswords}).
It achieves an accuracy of $0.800 (\pm 0.0616)$, significantly outperforming the previous state-of-the-art method, TOT evaluated with GPT-4, with an accuracy of $0.675$~\citep{yao2023tree}.
For a fair comparison, we also evaluate the TOT implementation by~\citet{yao2023tree} with GPT-4-Turbo.
It yields an accuracy of $0.668$.

\subsection{HumanEval}
\noindent \textbf{Motivation:}
In the previous experiments on MMLU (math problems) and Mini Crosswords (open-ended puzzles) we have validated the utility of optimizing graph edges.
In this section, we test the HumanEval dataset~\citep{chen2021evaluating}, which is known to be sensitive to prompt design.
Previous research involved manually crafting prompts~\citep{shinn2023reflexion,hong2023metagpt} and achieved impressive performance.
In contrast, here, we explore how node-based optimization can simplify this process.

\noindent \textbf{Analysis:}
In this section, we optimize the prompts of a ReAct-style~\citep{yao2022react} agent.
The agent first generates a Python program in response to a given question.
If the generated program passes all test cases included in the problem statement, then the program is returned.
Otherwise, the agent regenerates a program based on the execution feedback.
We optimize the prompts by adding input-output pairs selected by assessing their effectiveness as demonstration examples, particularly in improving the node operation applied to the node's ten most recent inputs.
To evaluate a node operation, we determine if the generated program successfully solves the unit tests provided in the input problem statement.
For more details on the node optimizer, please refer to Appendix \ref{app:humaneval_opt}.

\begin{figure}[t!]
    \centering
\includegraphics[width=0.5\textwidth]{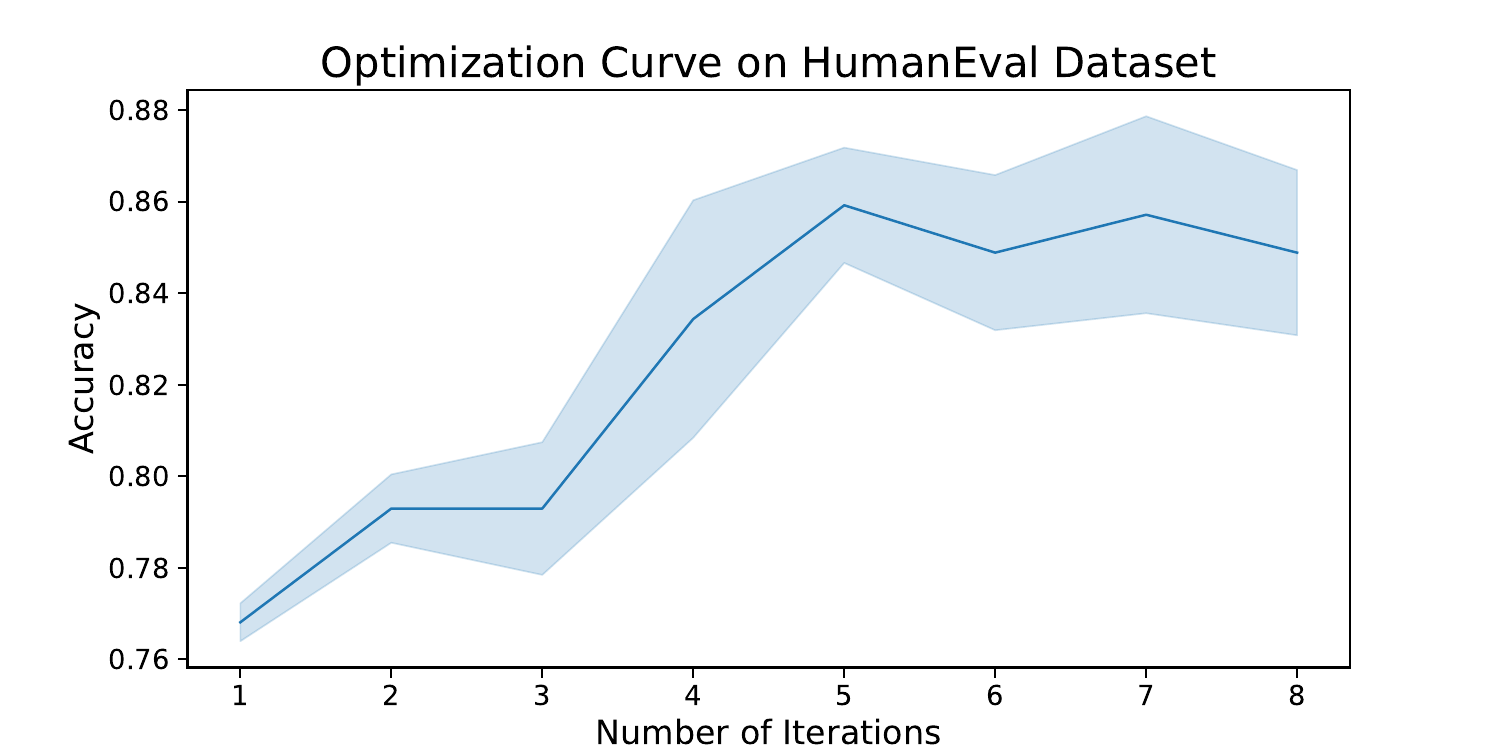}
    \caption{\textbf{Optimization curve on HumanEval.} Accuracy as a function of the number of iterations. We observe significant improvements during the first five iterations. The results, including the mean and standard errors, are based on three repeated experiments.}
    \label{fig:HumanEval}
\end{figure}

Figure \ref{fig:HumanEval} shows the optimization results of the first eight iterations.
After each iteration, the optimized solution is evaluated on the entire dataset.
The mean and standard errors are presented for three repeated runs.
We observe that the accuracy increases during the first five iterations.
We also test our method in an online learning setting, continuously optimizing and evaluating without restarting.
An improvement is observed from 0.76 (no optimization) to $0.88 (\pm 0.007)$ (mean and standard error).

 \begin{figure}[h!]
\centering
\vspace{-5pt}
\includegraphics[width=0.37\textwidth]{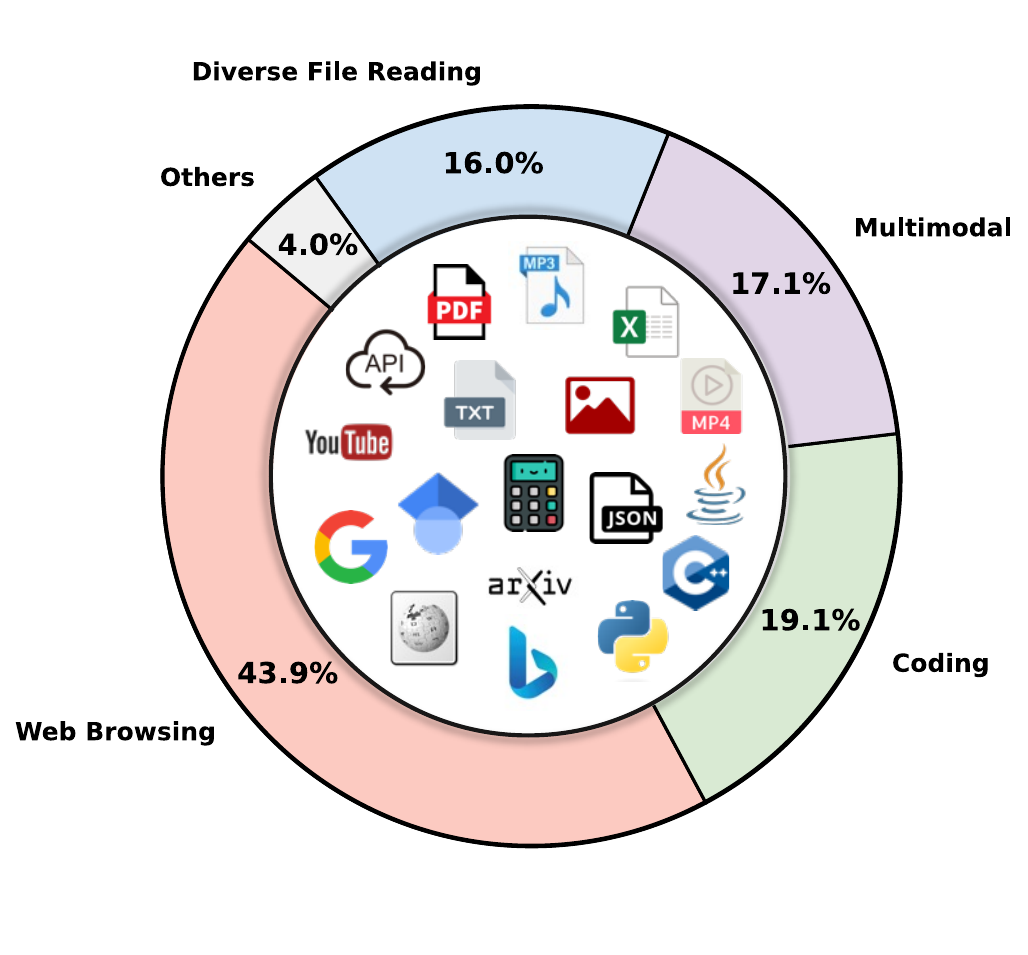}
\vspace{-25pt}
\caption{
\textbf{Solving a wide range of tasks requires many different tools.}
The GAIA benchmark~\citep{mialon2023gaia} tests for many of these capabilities by including questions that require several of these tools for successful completion.
}
\vspace{-10pt}
\label{fig:gaia_ratio}
\end{figure}

\subsection{GAIA}
\label{sec:gaia}

\begin{table}[t!]
  \centering
  \caption{
  \textbf{Performance on the GAIA Benchmark~\citep{mialon2023gaia}.}
  Using our framework, we demonstrate significant improvements across several levels of difficulty.
  The `GPT-4 with plugins' baseline is less significant since it involves the manual selection of the appropriate tools per question.
  We report the mean and standard deviation across 5 runs.
  }
  \label{tab:gaia}
  
  \renewcommand\tabcolsep{2.9pt}
  \renewcommand\arraystretch{1.1}
  \footnotesize 
  \begin{tabular}{l|cccc} 
    \Xhline{1pt}
    \rowcolor{champagne} 
    \textbf{Method} & \textbf{Level 1} & \textbf{Level 2} & \textbf{Level 3} & \textbf{Avg.} \\
    \Xhline{1.2pt}
    GPT-3.5 & 7.55 & 4.65 & 0 & 4.85 \\
    \rowcolor{gray!10}GPT-4 & 15.09 & 2.33 & 0 & 6.06 \\
    GPT-4-Turbo & \underline{20.75} & \underline{5.81} & 0 & \underline{9.70} \\
    \rowcolor{gray!10} AutoGPT & 13.21 & 0 & \underline{3.85} & 4.85 \\
    \hline
    GPTSwarm 
    & \textbf{30.56}$_{\pm\text{3.25}}$
    & \textbf{20.93}$_{\pm\text{1.27}}$
    & \textbf{3.85}$_{\pm\text{2.43}}$
    & \textbf{18.45}
    \\
    \rowcolor{gray!10}
    {\textit{Improvement}}
    & \textcolor{green(pigment)}{47.3\%$\uparrow$}
    & \textcolor{green(pigment)}{260.2\%$\uparrow$}
    & \textcolor{green(pigment)}{0.0\%}
    & \textcolor{green(pigment)}{90.2\%$\uparrow$}
    \\
    \hline
    \textcolor{black!50}{GPT4 with Plugins*}  & \textcolor{black!50}{30.30} & \textcolor{black!50}{9.70} & \textcolor{black!50}{0} & \textcolor{black!50}{14.6} \\
    \Xhline{1.2pt}
  \end{tabular}
\end{table}

\noindent \textbf{Motivation:} 
GAIA is a benchmark specifically designed for testing the generality of AI assistants focusing on real-world questions~\citep{mialon2023gaia}.
Abilities required to answer GAIA questions include reasoning, multi-modality processing, web browsing, and other tool use.
Although conceptually straightforward for humans, these questions present significant challenges for current AI systems.

Using this benchmark, we evaluate the general applicability of our framework.
We construct swarms with multiple agents of the same type and employ self-consistency (a prompt-based majority vote) for the final decision~\cite{wang2022self}.
We also experimented with adding different types of agents to the swarm and using prompt-based best answer selection.
The results indicate that prompt-based self-consistency yields the best performance.
Note that these experiments are meant to demonstrate the generic capabilities of our modular framework and include neither edge-based nor node-level optimization, which is left for future work.

\begin{table*}[h!]
\centering
\caption{
\textbf{Ablations on the GAIA benchmark (Level 1 validation set)~\citep{mialon2023gaia}.} 
DA = DirectAnswer, GQ = GenerateQuery, WS = WebSearch, FA = FileAnalyzer, CA = CombinedAnswer.
`\textcolor{green(pigment)}{\Checkmark}' indicates the presence of a specific feature in the corresponding framework, `\textcolor{darksalmon}{\XSolidBrush}' its absence.
Each type of experiment is run five times to record the mean, standard deviation, and best run (marked as Best).
Self-Consistency describes prompt-based self-consistency~\citep{wang2022self};
Choose ``Best'' refers to the LLM's favorite answer among the different agents' answers.
All agents and swarms are implemented using our GPTSwarm framework.
}
\label{tab:gaia_abl}

\renewcommand\tabcolsep{8.35pt}
\renewcommand\arraystretch{1}
\footnotesize 
\begin{tabular}{lcccccc|ccc}
\Xhline{1.2pt}
\rowcolor{champagne} 
\textbf{Agent or Swarm} & \textbf{DA} & \textbf{GQ} & \textbf{WS} & \textbf{FA} & \textbf{CA} & \textbf{Decision Strategy} & \textbf{Accuracy} & \textbf{Best} & \textbf{Duration (s)} \\
\Xhline{1.2pt}
\textbf{(A)} Agent: IO                          & \textcolor{green(pigment)}{\Checkmark} & \textcolor{darksalmon}{\XSolidBrush} & \textcolor{darksalmon}{\XSolidBrush} & \textcolor{darksalmon}{\XSolidBrush} & \textcolor{darksalmon}{\XSolidBrush} & N/A & 16.60\textcolor{green(pigment)}{$\pm$\text{3.02}} & 20.75\% & $\sim$13.37 \\
\rowcolor{gray!10}\textbf{(B)}  Agent: COT$_\text{web}$                    & \textcolor{darksalmon}{\XSolidBrush} & \textcolor{green(pigment)}{\Checkmark} & \textcolor{green(pigment)}{\Checkmark} & \textcolor{darksalmon}{\XSolidBrush} &\textcolor{green(pigment)}{\Checkmark} & N/A & 18.87\textcolor{green(pigment)}{$\pm$\text{2.67}} & 22.64\% & $\sim$60.90 \\
\textbf{(C)}  Agent: COT$_\text{FA}$                      & \textcolor{darksalmon}{\XSolidBrush} & \textcolor{green(pigment)}{\Checkmark} & \textcolor{darksalmon}{\XSolidBrush} & \textcolor{green(pigment)}{\Checkmark} & \textcolor{green(pigment)}{\Checkmark} & N/A & 25.28\textcolor{green(pigment)}{$\pm$\text{3.50}} & 30.18\%  & $\sim$56.42 \\
\rowcolor{gray!10}\textbf{(D)}  Agent: TOT                         & \textcolor{darksalmon}{\XSolidBrush} & \textcolor{green(pigment)}{\Checkmark} & \textcolor{green(pigment)}{\Checkmark} & \textcolor{green(pigment)}{\Checkmark} & \textcolor{green(pigment)}{\Checkmark} & N/A & 25.66\textcolor{green(pigment)}{$\pm$\text{3.50}} & 30.18\% & $\sim$71.31 \\
\hline
\textbf{(E)} Swarm$_\text{(3$\times$IO)}$  & \textcolor{green(pigment)}{\Checkmark} & \textcolor{darksalmon}{\XSolidBrush} & \textcolor{darksalmon}{\XSolidBrush} & \textcolor{darksalmon}{\XSolidBrush} & \textcolor{darksalmon}{\XSolidBrush}& Choose ``Best''  & 15.85\textcolor{green(pigment)}{$\pm$\text{0.92}} & 18.87\% & $\sim$45.65 \\
\textbf{(F)} Swarm$_\text{(3$\times$COT)}$       & \textcolor{darksalmon}{\XSolidBrush} & \textcolor{green(pigment)}{\Checkmark} & \textcolor{green(pigment)}{\Checkmark} & \textcolor{darksalmon}{\XSolidBrush} &\textcolor{green(pigment)}{\Checkmark} & Choose ``Best'' & 27.17\textcolor{green(pigment)}{$\pm$\text{3.29}} & 32.08\% & $\sim$152.89 \\
\textbf{(G)} Swarm$_\text{(3$\times$TOT)}$  & \textcolor{darksalmon}{\XSolidBrush} & \textcolor{green(pigment)}{\Checkmark} & \textcolor{green(pigment)}{\Checkmark} & \textcolor{green(pigment)}{\Checkmark} & \textcolor{green(pigment)}{\Checkmark} & Choose ``Best'' & 30.18\textcolor{green(pigment)}{$\pm$\text{4.30}} & 35.85\% & $\sim$198.50 \\
\rowcolor{gray!10}\textbf{(H)} Swarm$_\text{(3$\times$IO)}$  & \textcolor{green(pigment)}{\Checkmark} & \textcolor{darksalmon}{\XSolidBrush} & \textcolor{darksalmon}{\XSolidBrush} & \textcolor{darksalmon}{\XSolidBrush} & \textcolor{darksalmon}{\XSolidBrush} & Self-Consistency & 18.11\textcolor{green(pigment)}{$\pm$\text{3.07}} & 22.64\% & $\sim$45.70 \\
\rowcolor{gray!10}\textbf{(I)} Swarm$_\text{(3$\times$COT)}$  & \textcolor{darksalmon}{\XSolidBrush} & \textcolor{green(pigment)}{\Checkmark} & \textcolor{green(pigment)}{\Checkmark} & \textcolor{darksalmon}{\XSolidBrush} &\textcolor{green(pigment)}{\Checkmark} & Self-Consistency & 27.17\textcolor{green(pigment)}{$\pm$\text{4.06}} & 32.08\% & $\sim$150.26 \\
\rowcolor{gray!10}\textbf{(J)} Swarm$_\text{(3$\times$TOT)}$ & \textcolor{darksalmon}{\XSolidBrush} & \textcolor{green(pigment)}{\Checkmark} & \textcolor{green(pigment)}{\Checkmark} & \textcolor{green(pigment)}{\Checkmark} & \textcolor{green(pigment)}{\Checkmark} & Self-Consistency & 28.30\textcolor{green(pigment)}{$\pm$\text{3.38}} & 32.08\% & $\sim$181.15 \\
\hline
\textbf{(K)} Swarm$_\text{(5$\times$TOT)}$ & \textcolor{darksalmon}{\XSolidBrush} & \textcolor{green(pigment)}{\Checkmark} & \textcolor{green(pigment)}{\Checkmark} & \textcolor{green(pigment)}{\Checkmark} & \textcolor{green(pigment)}{\Checkmark} & Self-Consistency & 29.06\textcolor{green(pigment)}{$\pm$\text{2.56}} & 32.08\% & $\sim$291.07 \\
\rowcolor{gray!10}\textbf{(L)} Swarm$_\text{(7$\times$TOT)}$ & \textcolor{darksalmon}{\XSolidBrush} & \textcolor{green(pigment)}{\Checkmark} & \textcolor{green(pigment)}{\Checkmark} & \textcolor{green(pigment)}{\Checkmark} & \textcolor{green(pigment)}{\Checkmark} & Self-Consistency & 30.56\textcolor{green(pigment)}{$\pm$\text{3.25}} & 35.85\% & $\sim$414.89 \\
\hline
\textcolor{black!50}{\textbf{(M)} Human} & \textcolor{black!30}{-} & \textcolor{black!50}{-} & \textcolor{blue!50}{-}& \textcolor{black!50}{-} & \textcolor{black!50}{-} & \textcolor{black!50}{-} & \textcolor{black!50}{94\%} & \textcolor{black!50}{-} & \textcolor{black!50}{$\sim$422.26}\\
\Xhline{1.2pt}
\end{tabular}
\end{table*}

\noindent \textbf{Analysis:} 
Table \ref{tab:gaia} shows the results of our swarm with seven TOT agents and the self-consistency strategy for the final decision.
We compare the performance of the GPT-Series~\citep{achiam2023gpt} with plugins and AutoGPT~\citep{autogpt_github} performance as reported by \citet{mialon2023gaia}. 
Our methods significantly outperform these baselines.

Table \ref{tab:gaia_abl} presents a more comprehensive set of results.
We experiment with varying numbers of agents and different node operations, such as different tool uses.
Our observations indicate that the time requirement of a swarm grows approximately linearly with the number of agents. 
Despite the increased computational time, incorporating more agents notably improves the overall performance of the system.
We also found that a greater variety of node operations leads to better performance. 
As illustrated in \Cref{fig:gaia_ratio}, web browsing is required for 43.9\% of the tasks.
Our current implementation accesses the Internet by only downloading materials directly from the URLs provided in the problem statement or querying a Google search\footnote{We use SearchApi \href{https://www.searchapi.io/}{(https://www.searchapi.io/)} in the experiments} without further website navigation.
Therefore, we believe that enhancing web capabilities would further increase performance significantly.

\section{Related Work}

\subsection{LLM-based Autonomous Agents}

Current works on LLM-based autonomous agents or language agents vary in focus. 
Methods such as Chain of Thought~\citep{wei2022chain}, ReAct~\citep{yao2022react}, Reflexion~\citep{shinn2023reflexion}, Tree of Thought (ToT) improve prompt strategies and structure to improve reasoning capabilities, while others such as exchange-of-thoughts (EOT)~\cite{yin2023exchange} focus on the benefits of various communication paradigms.
Single LLM agent frameworks such as AutoGPT~\citep{gravitas2023auto}, LangChain~\citep{Chase_LangChain_2022}, LlamaIndex~\citep{Liu_LlamaIndex_2022}, and XAgent~\citep{xagent2023} showcase problem solving through various external functions and tools. 
In the space of LLM-based multiagent systems~\citep{xie2023openagents,chen2023autoagents,chen2023agentverse}, NLSOMs~\citep{zhuge2023mindstorms} employ various social structures for task-specific applications (inspired by SOMs~\citep{minsky1988society}), without exploring optimization over the social structure of agents.
CAMEL~\citep{li2023camel}, Generalist Agents~\citep{park2023generative}, ChatDev~\citep{qian2023communicative}, and AutoGen~\citep{wu2023autogen} focus on role-based communication, but struggle with hallucinations.
MetaGPT~\citep{hong2023metagpt} introduces standard operating procedures for better role definition and communication, making the collaboration between agents more effective.
In contrast to these frameworks, we automatically optimize nodes and edges in a self-organizing society of agents. 

\subsection{Language Agents with Graphs}

\citet{besta2023graph} introduced LLM-based problem-solving with graphs; however, the approach only encompasses LLM prompting schemes without modeling other fundamental capabilities of language agents, such as use of external tools.
LangGraph~\citep{langgraph}, on the other hand, is a concurrent open-source framework that focuses on building multi-actor state LLM applications through possibly cyclical operations.
However, its practical applicability has not yet been systematically studied.
Unlike previous studies, our approach emphasizes the development of hierarchical intelligence, as discussed by~\citet{minsky1988society} and~\citet{kennedy2006swarm}, through the construction of agent graphs and the composition of multiple graphs into swarms.
Crucially, the graph representation facilitates automatic optimization on two levels.
First, at the node level, since the majority of nodes in the graph involve prompting an LLM, prompt optimization methods can be employed.
Second, at the edge level, we demonstrate the application of the REINFORCE algorithm~\citep{williams1992simple} to optimize the potential connections between nodes.

\subsection{Optimizing LLM Inference and Self-Improvement}
Much of deep learning research is concerned with tuning the learning algorithms, architectures, hyper-parameters, and other aspects of the learning pipeline~\citep{schmidhuber2015deep,yan2015deep}.
Meta-learning attempts to automate large parts of that process~\citep{schmidhuber1987evolutionary,elsken2019neural,kirsch2020meta}.
Similarly, recently, a lot of research and engineering has gone into the prompting and structuring of LLM inference to make better use of LLMs and build better agents.
Due to the ability of LLMs to learn in context~\citep{brown2020language,kirsch2022general}, one can view this process as configuring learning algorithms.
The optimization of the inference structure and the prompts can then be viewed as meta-learning in LLMs.

In the realm of prompt optimization, OPRO~\citep{yang2023large} generates better prompts through iterative LLM queries using prior solutions and their performance.
PromptBreeder~\citep{fernando2023promptbreeder} implements a mechanism that evolves and self-improves task-specific and meta-prompts through mutation and LLM prompting.
Related to these works, we self-improve future prompts by prompting LLMs.
Similarly to our work, DSPy~\citep{khattab2023dspy} implements LLM pipelines as computational graphs with modular LLM queries as nodes, parameterized by prompts and neural network weights. 
It proposes a two-stage process to optimize the parameters of these nodes.
Initially, it generates a set of candidate solutions for each node.
Subsequently, it optimizes across the Cartesian product of these candidate solution sets, aiming to identify an effective combination of parameters for the entire graph.
To address the combinatorial optimization challenge raised in DSPy, we propose an iterative optimization process.
By virtue of decomposing a solution into nodes with expected functions, at each iteration, we improve each node individually, conditioned on the execution history of the graph with the current prompts of each node.

Regarding the optimization of the inference structure, DyLAN~\citep{liu2023dynamic} uses a fixed heuristic to improve the collaboration of LLM agents by selecting agents and determining the number of communication rounds.
In line with previous ideas on self-referential learning~\citep{schmidhuber1993self,irie2022modern,kirsch2022eliminating}, STOP~\citep{zelikman2023self} optimizes both the prompts and the inference structure together by introducing an initial improver program that is applied to itself to iteratively improve its performance.
In our work, we optimize the inference structure by employing RL techniques applied to the potential edges of a given graph.

\section{Conclusion}
This paper introduces GPTSwarm, an open-source framework that constructs language agents from graphs and agent societies from graph compositions.
This approach allows for the easy implementation of existing methods from basic node operations and enables automatic optimization of the graph in the form of node-level improvement and edge-level REINFORCE optimization.
Our experiments demonstrate the advantages of our language agent graphs and automatic optimization on several benchmarks.

\section*{Impact Statement}
This paper presents work whose goal is to advance the field of Machine Learning.
The societal consequences of our work are multifaceted.
On the one hand, it could lead to significant advancements in the efficiency and effectiveness of machine learning systems.
On the other hand, the increased capability and automation of LLM agents might raise ethical and employment concerns.
As AI systems become more autonomous and powerful, it is crucial to consider their impact on job displacement and the importance of implementing safeguards to prevent biased or unethical AI behaviors.
Furthermore, the potential for misuse of advanced AI technologies requires rigorous oversight and the development of ethical guidelines to ensure that these technologies are used responsibly and for the benefit of society as a whole.

\section*{Author Contributions}
Mingchen initiated the project and conceived the initial idea, led the development of the codebase, conducted GAIA \& HumanEval experiments, drafted the initial manuscript, and created most of the visualizations.
Wenyi discussed the core ideas with Mingchen, contributed to the codebase, conducted Mini CrossWords \& HumanEval experiments, and drafted the initial manuscript.
Louis reviewed and polished the paper, extensively rewrote the introduction, and coordinated team meetings.
Francesco reviewed and polished the paper, significantly revising the methods section.
Furthermore, Louis and Francesco discussed and formalized various techniques for graph optimization.
As the senior engineering lead, Dmitrii advised and made significant revisions to the codebase, conducted the MMLU experiments, and contributed to the visualizations.
Juergen, as mentor and advisor, offered guidance and support throughout the project's progression.

\section*{Acknowledgements}
This work was supported by the SDAIA-KAUST Center of Excellence in Data Science and Artificial Intelligence (SDAIA-KAUST AI).
It was further supported by the European Research Council (ERC, Advanced Grant Number 742870) and the Swiss National Science Foundation (SNF, Grant Number 200021 192356).

\bibliography{main}
\bibliographystyle{icml2024}

\newpage
\appendix
\onecolumn
\section{The GPTSwarm Framework}\label{apendix:gptswarm_engineering}

\subsection{The Vision}
Many recent language agents are described as compositions of components of different functionalities~\citep{wang2023survey}. A popular tweet states: ``agent = LLM + memory + planning skills + tool use.''\footnote{\href{https://twitter.com/lilianweng/status/1673535600690102273}{https://twitter.com/lilianweng/status/1673535600690102273}}
Such additive formulations highlight individual components, but fail to address the essential aspect of component {\em integration}. GPTswarm's computational graph formulation, however, precisely focuses on {\em integration} through edge optimization, to learn improved recommendations of agent orchestration and precise agent routing. This will become increasingly relevant as swarm size increases to millions or billions of agents.

\subsection{Class Diagram}
\begin{figure*}[!htbp]
    \centering
    \includegraphics[width=0.98\textwidth]{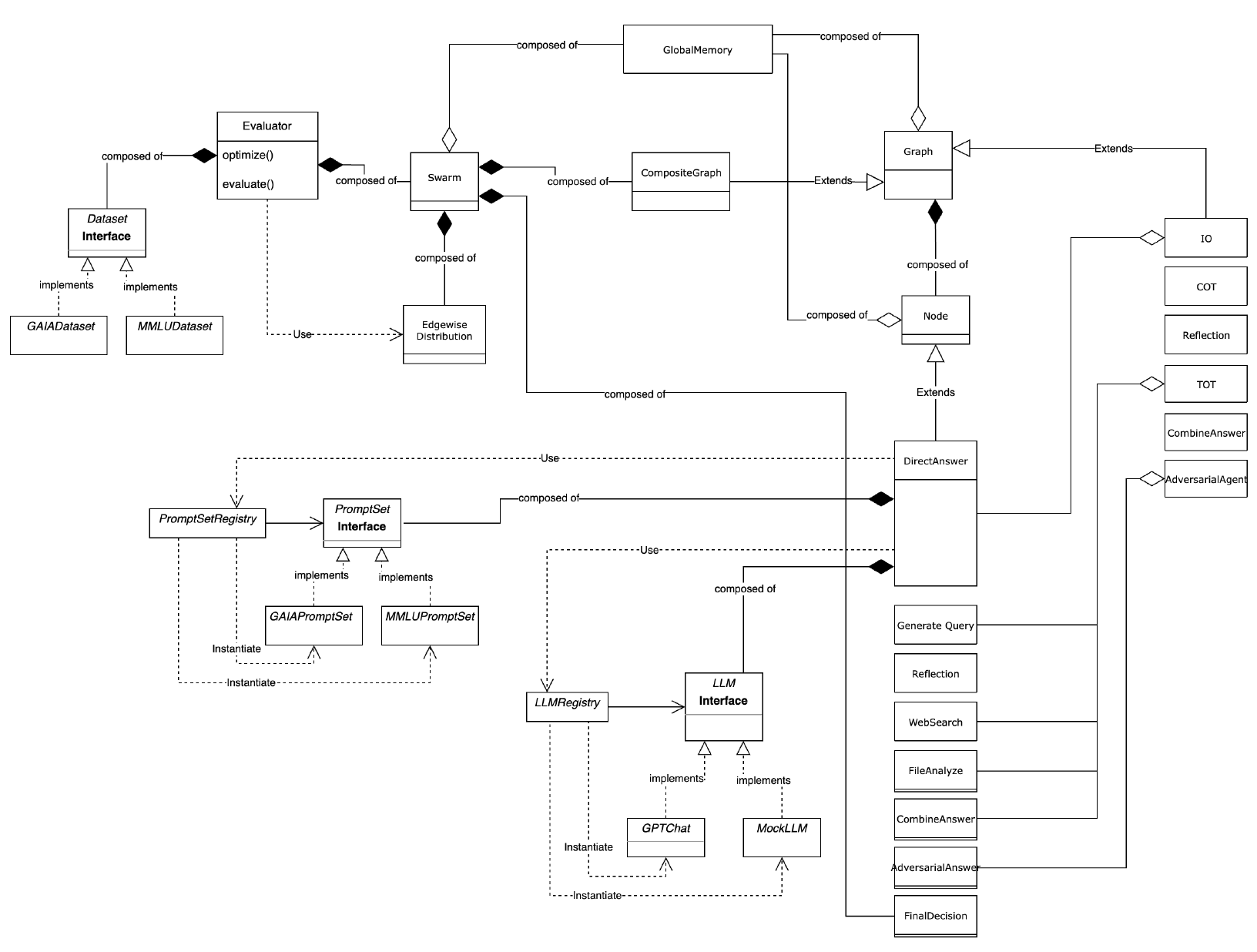}
    \caption{The class diagram of the GPTSwarm framework.}
    \label{fig:class_diagram}
\end{figure*}

The GPTSwarm framework is developed using Python and PyTorch.
Its class diagram is illustrated in \Cref{fig:class_diagram}.

On the graph level, Node, Graph, and CompositeGraph are directly implemented as classes.
Graph edges are implicitly stored as an adjacency list within each Node.
Functionally, the framework distinguishes between Agents and Operations through various classes, such as DirectAnswer and WebSearch for operations, and IO and TOT for agents.
To encapsulate the abstraction of an external LLM, we introduce an interface named after it.
The primary implementation of this interface is a lightweight wrapper around the OpenAI API.
To facilitate dataset integration for optimization and evaluation, we provide implementations for two interfaces: Dataset and PromptSet.
The Dataset interface is designed to load benchmark datasets like GAIA and MMLU, while the PromptSet customizes node behavior for a specific Dataset.
The Evaluator class manages the optimization processes for both edges and nodes.

The framework is highly customizable, allowing users to add more LLM backends, Dataset and PromptSet combinations, Agents, and Nodes as needed.
Additionally, the framework makes extensive use of asynchronous computations for task parallelism, leveraging Python's async-await syntax.

\newpage
\section{Swarm examples}

\begin{figure}[!h]
    \centering
    \includegraphics[width=0.8\textwidth]{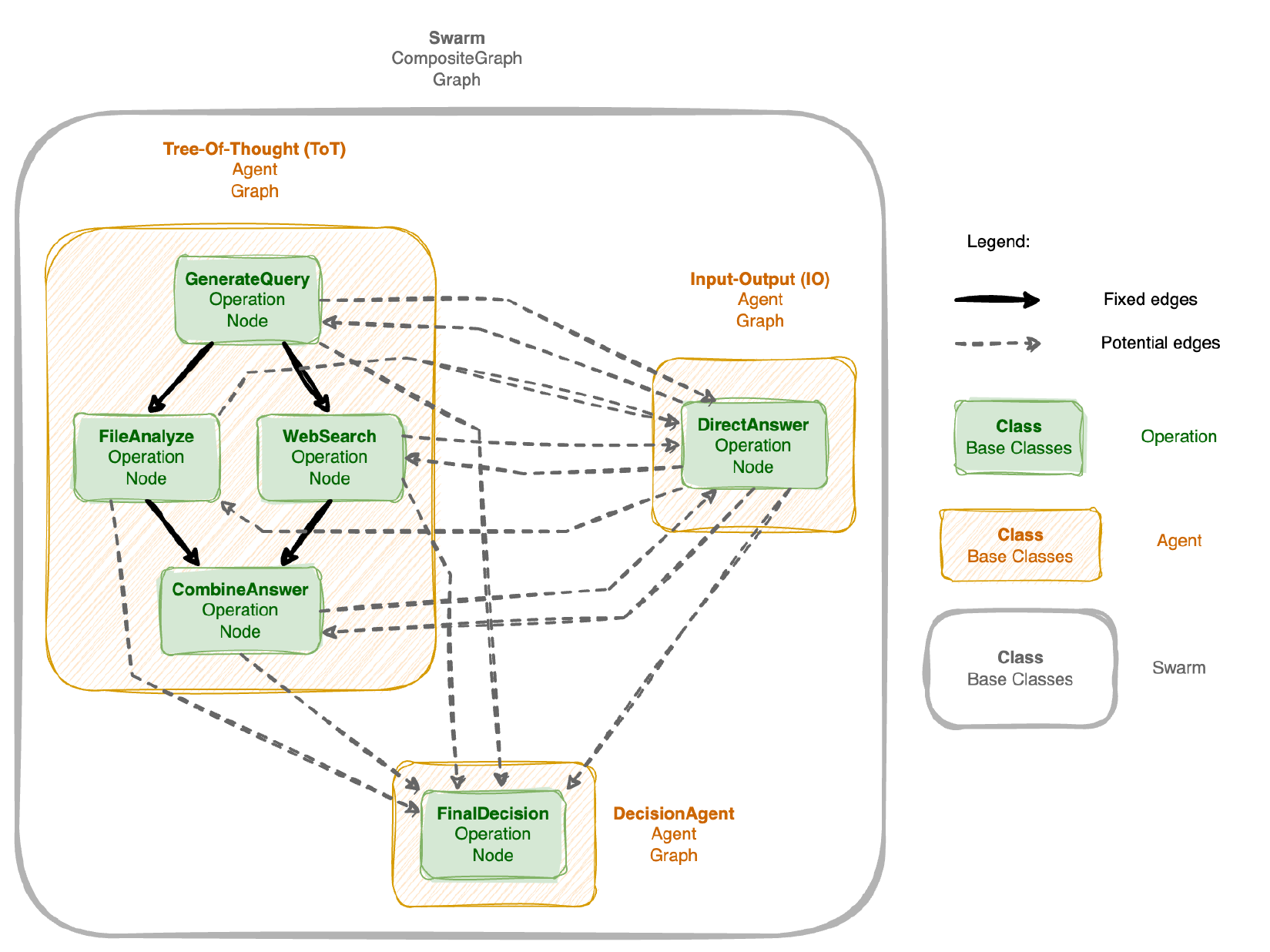}
    \caption{A simple example of a swarm consisting of one Tree-of-Thought, one Input-Output, and the Decision agent.
    }
    \label{fig:example_tot_io}
\end{figure}

To facilitate understanding of the concepts presented in this study, we are showing a simple example of a swarm consisting of 3 agents: Tree-of-Thought, Input-Output, and Decision Agents in \Cref{fig:example_tot_io}.

\newpage
\section{More Visualizations}

\begin{figure}[!h]
    \centering
    \vspace{15pt}
    \includegraphics[width=1\textwidth]{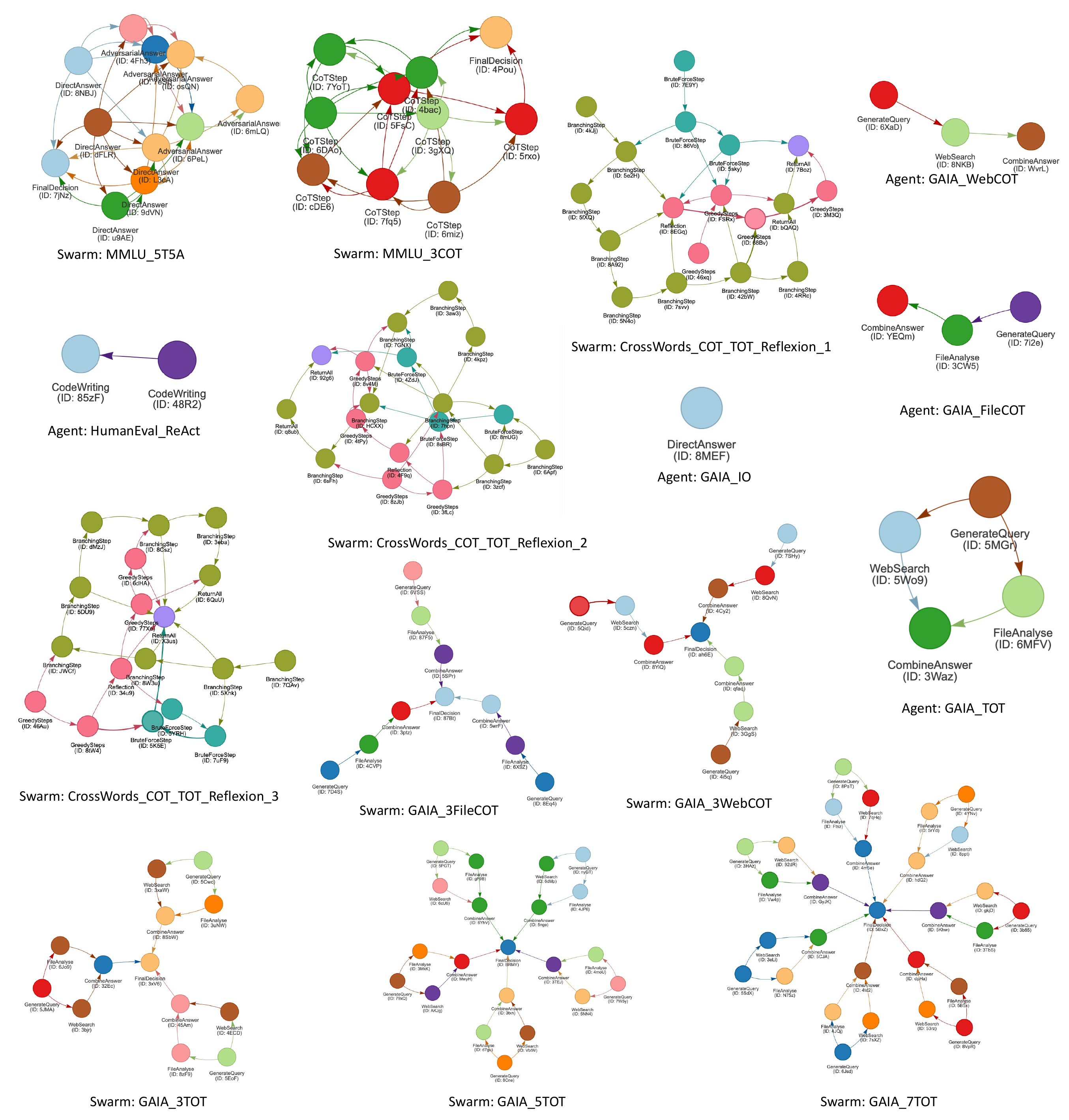}
    \caption{Different agents or swarms implemented by GPTSwarm.
    }
    \label{fig:all_swarm_examples}
\end{figure}

\newpage

\section{Additional Experiments}
\subsection{Comparing Our Method with \emph{Multiagent Debate} and \emph{DYLAN} on MMLU}
Our MMLU experimental setup can be viewed as a multi-agent optimization problem with an equal number of truthful and adversarial agents in the system.
We compare our approach to two baselines from the literature that naturally fit this problem:
Multiagent Debate~\citep{du2023improving} and DyLAN~\citep{liu2023dynamic}.

For the Multiagent Debate baseline, we directly use the original setting with three truthful agents and three adversarial agents.
For DyLAN, we adopt the original implementation~\citep{liu2023dynamic}, adding system prompts "You are a knowledgeable expert in question answering." for truthful agents, and "Pretend you are a non-expert in question answering. You can provide wrong answers to the questions." for adversarial agents.
DyLAN performs optimization by scoring each agent and pruning a certain number of agents with the lowest scores.
Since DyLAN has not fixed the number of pruned agents, we report the average performance in multiple settings, from 1 to 5 agents pruned.

The results in Table \ref{tab:MMLU} indicate that the Multiagent Debate is less effective in this setting.
DyLAN achieves a $0.0065$ improvement in accuracy compared to our method.
We suspect that this improvement is due to DyLAN's more complex debate scheme and prompts.
This design also contributes to DyLAN's larger computational cost.
\begin{table*}[h!]
  \centering
  \caption{\textbf{Results of the Multiagent Debate, DyLAN, and our method on MMLU.}
  We report the performance and computational cost of these methods applied to an LLM-based multiagent system with adversaries.
  Computational cost for optimization and inference is presented separately except for the Multiagent Debate, where there is no explicit separation.
  DyLAN is reported as an average over five different choices for the number of pruned agents.}
  \label{tab:MMLU}

  \renewcommand\tabcolsep{12pt}
  \renewcommand\arraystretch{1.2}
  \footnotesize
  \begin{tabular}{l|ccccc}
    \Xhline{1.2pt}
    \rowcolor{champagne}
    \textbf{Methods} & \textbf{Cost (USD)} & \textbf{\# Prompt Tokens} & \textbf{\# Completion Tokens} & \textbf{Time (h)} & \textbf{Accuracy} 
    \\
    \Xhline{1.2pt}

    Multiagent Debate & 32.8 & 1,689,960 & 530,005 & 8.36 & 0.5751 
    \\
    \rowcolor{gray!10} DyLAN optimization & 105.93 & 5,671,276 & 1,640,566 & 25.4 & - 
    \\
    DyLAN inference & 14.99 & 628,009 & 290,472 & 4.75 & 0.8366 
    \\
    \rowcolor{gray!10} GPTSwarm optimization & 5.32 & 361,812 & 56,770 & 0.9 & - 
    \\
    GPTSwarm inference & 1.82 & 113,233 & 22,923 & 0.31 & 0.8301 
    \\
    \Xhline{1.2pt}
  \end{tabular}
\end{table*}
\subsection{Applying Node Optimization after Edge Optimization}
As described in Section 3.2, we first perform ten REINFORCE steps to optimize the edges of a three-agent swarm, resulting in a $0.575(\pm0.0275)$ accuracy.
Based on this optimized edge distribution we further apply our node optimization algorithm (Algorithm \ref{alg:node_opt}).
For each node $n$, $p_n$ corresponds to the initial prompt, $h_n$ consists of the input-output pairs obtained by executing the graphs on the 20 Mini Crosswords problems, and d is ignored.
For each node $n$, the improver $I$ either chooses a demonstration example from $h_n$ to include as a part of $n$’s prompt or keeps the prompt unchanged.
This choice is implemented using the upper confidence bound algorithm UCB1~\citep{auer2002finite}, over a hundred iterations.
Preference is given to pairs (or the choice of no demonstration example) that help the node accurately fill in more words on its previous input.
Applying this node optimization after edge optimization improves accuracy to $0.668 (\pm 0.0060)$. 

\section{Experimental Details}
\label{app:experiment}
In our experiments involving multiple agents, we incorporate an additional virtual agent, represented by a single node, to serve as a final decision aggregator.
This node is designated as the output node for the composite graph, and its specific implementation varies between different experiments.
Common implementations for this node include employing a majority vote and a self-consistency strategy for decision-making.
Unless explicitly stated, communication between agents within a composite graph does not include this virtual agent.
Additionally, in all our experiments, the potential edge set of a composite graph is defined as all possible node pairs, provided that the nodes in each pair originate from different agents.
We exclude any edges that would connect the output node of a composite graph to other nodes.
Moreover, we employ Adam~\citep{kingma2014adam} optimizer with parameters $\beta_1 = 0.9, \beta_2 = 0.999$ and a variable learning rate in place of the vanilla stochastic gradient accent method described in Alg. \ref{alg:reinforce}. Finally, we use the version gpt-4-1106-preview and gpt-3.5-turbo-1106 for LLMs. For the vision-language model utilized in the GAIA experiments, we employed the gpt-4-1106-vision-preview version of GPT-4-Turbo.

\subsection{MMLU}
\label{sec:mmlu_appendix}
\begin{figure}[h!]
\centering
\vspace{8pt}
\includegraphics[width=0.55\textwidth]{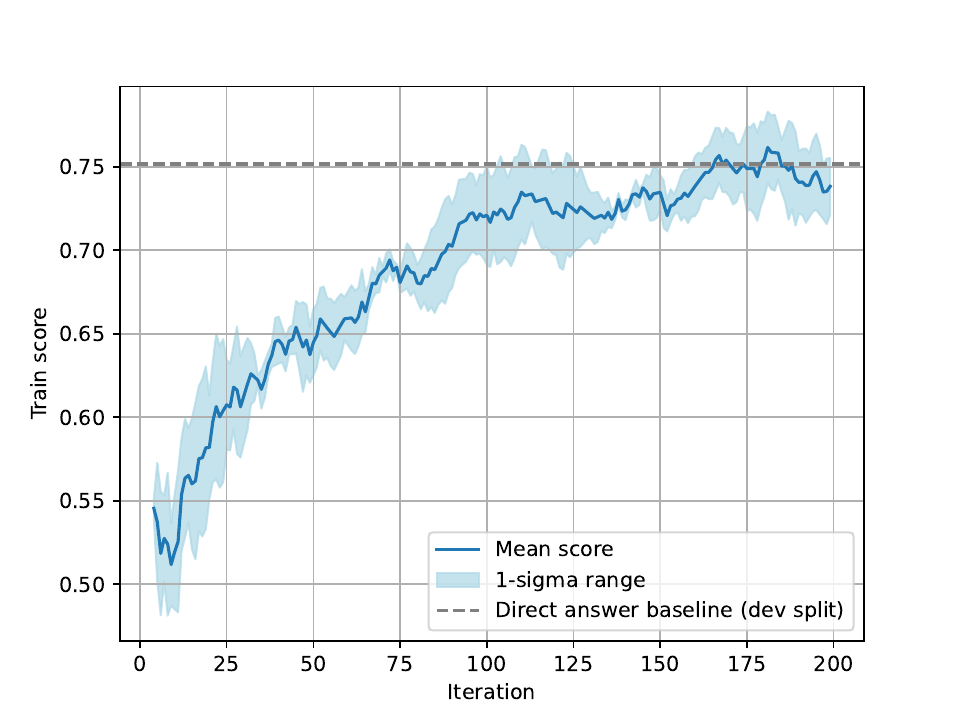}
\vspace{-15pt}
\caption{
\textbf{The training score during the optimization of the adversarial swarm (3T3A) on MMLU.}
We apply smoothing with an unbiased exponential moving average and the smoothness factor of $0.97$.
}
\label{fig:adv_training}
\end{figure}

We provide adversarial robustness experiments in \Cref{tab:adv_stats}.
The convergence of the train utility to the baseline for the 3T3A experiment is shown in \Cref{fig:adv_training}. The first 10\% of the validation dataset comprises 153 questions.

\begin{table*}[h!]
  \centering
  \caption{\textbf{Stats for the adversarial experiments.} \#Nodes means the number of nodes in the swarm excluding the final decision node. \#Potential edges is the total number of edges that are optimized and potentially realized. The optimization time is measured as the wall clock time. \#LLM inferences is the total number of LLM queries made during the optimization cycle when graph pruning is turned off.}
  \label{tab:adv_stats}
  
  \renewcommand\tabcolsep{20pt}
  \renewcommand\arraystretch{1.2}
  \footnotesize 
  \begin{tabular}{l|ccc} 
    \Xhline{1.2pt}
    \rowcolor{champagne}
    \textbf{Swarm configuration} & \textbf{\#Nodes} & \textbf{\#Potential edges} & \textbf{Optimization time, mins}  
    \\
    \Xhline{1.2pt}

    1 Trustful Agent + 1 Adversarial Agent & 2 & 4 & 9 
    \\
    \rowcolor{gray!10}3 Trustful Agents + 3 Adversarial Agents & 6 & 36 & 23 
    \\
    5 Trustful Agents + 5 Adversarial Agents & 10 & 100 & 58 
    \\
    \rowcolor{gray!10} 7 Trustful Agents + 7 Adversarial Agents & 14 & 196 & 95 
    \\
    \Xhline{1.2pt}
  \end{tabular}
\end{table*}

\subsubsection{Hyper-parameters \& Prompts}
We use the Adam optimizer with a learning rate of $0.1$ to update the logit parameters associated with each potential edge.
The prompts that have been used for the adversarial robustness experiments are collected in \Cref{tab:mmlu_prompts}.

\begin{table*}[h!]
  \centering
  \caption{\textbf{Prompts for the Adversarial experiments on MMLU.}}
  \label{tab:mmlu_prompts}
  
  \renewcommand\tabcolsep{10pt}
  \renewcommand\arraystretch{1.3}
  \footnotesize 
  \begin{tabular}{|p{0.2\linewidth}|p{0.7\linewidth}|} 
    \Xhline{1.2pt}
    \rowcolor{champagne} 
    \textbf{Prompt purpose} & \textbf{Prompt} \\
    \Xhline{1.2pt}
        System prompt & You are a knowledgeable expert in question answering.
        I will ask you a question.
        I will also give you 4 answers enumerated as A, B, C and D.
        Only one answer out of the offered 4 is correct.
        You must choose the correct answer to the question.
        Your response must be one of the 4 letters: A, B, C or D,
        corresponding to the correct answer.
        Only one letter (A, B, C or D) is allowed in your answer. \\
    \rowcolor{gray!10}Direct answer template & \{question\} \\
    \Xhline{1.2pt}
    Adversarial answer template & Answer a lie to the following question: 
    \{question\}. \\
    \Xhline{1.2pt}
    Question template & \{open-ended question\}. Option A: \{option A\}, Option B: \{option B\}, Option C: \{option C\}, Option D: \{option D\}. \\
    \Xhline{1.2pt}
  \end{tabular}
\end{table*}

\newpage
\subsubsection{Adversarial swarm optimization case study}
\label{apendix:adv_case}

\begin{figure}[h!]
\centering

\begin{subfigure}{0.37\textwidth}
    \includegraphics[width=\textwidth]{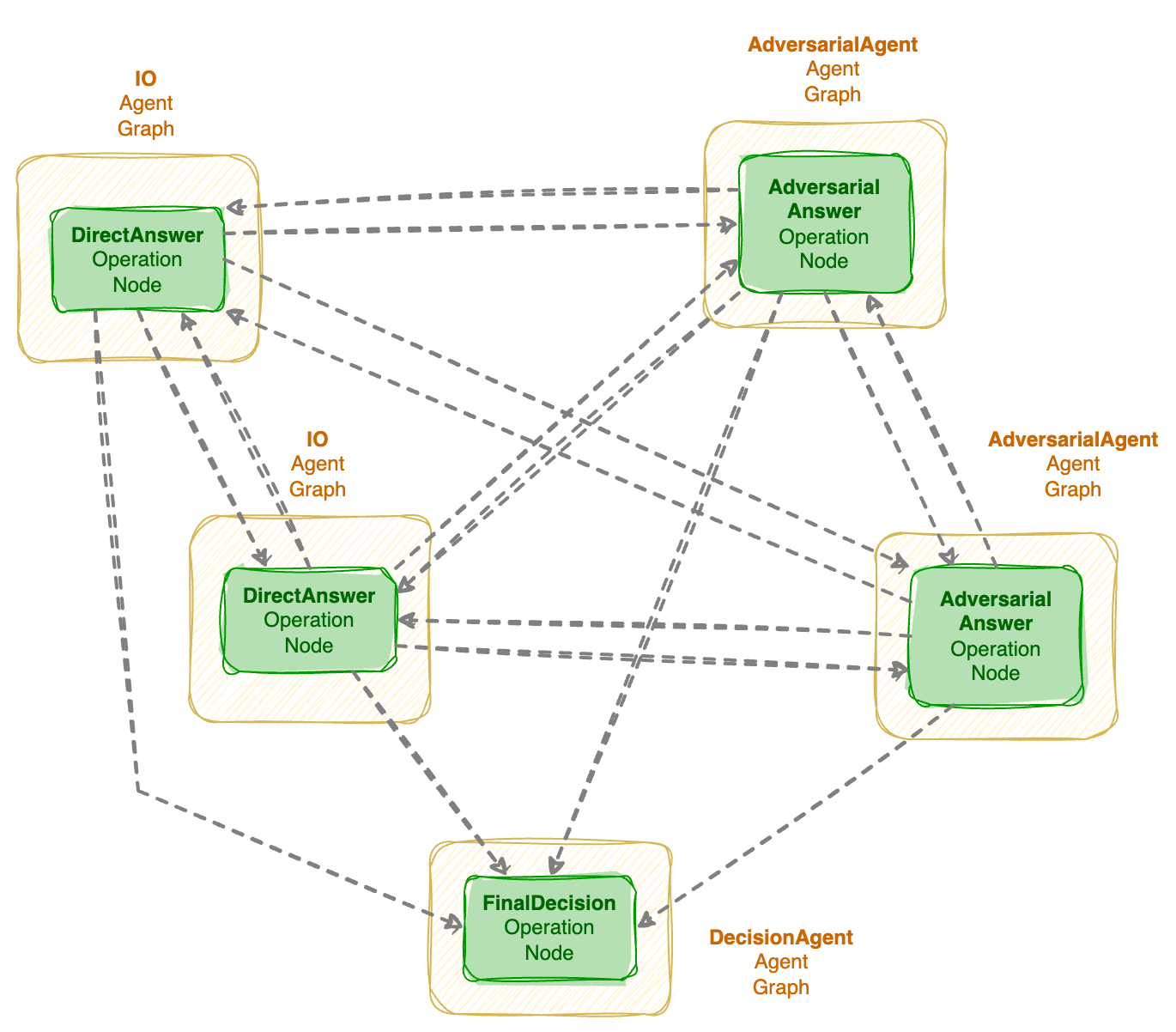}
    \caption{An non-optimized swarm with 2 truthful agents and 2 adversarial agents. Dotted arrows depict potential edges.}
    \label{fig:before_opt}
\end{subfigure}
\hfill

\begin{subfigure}{0.37\textwidth}
\vspace{20pt}
    \includegraphics[width=\textwidth]{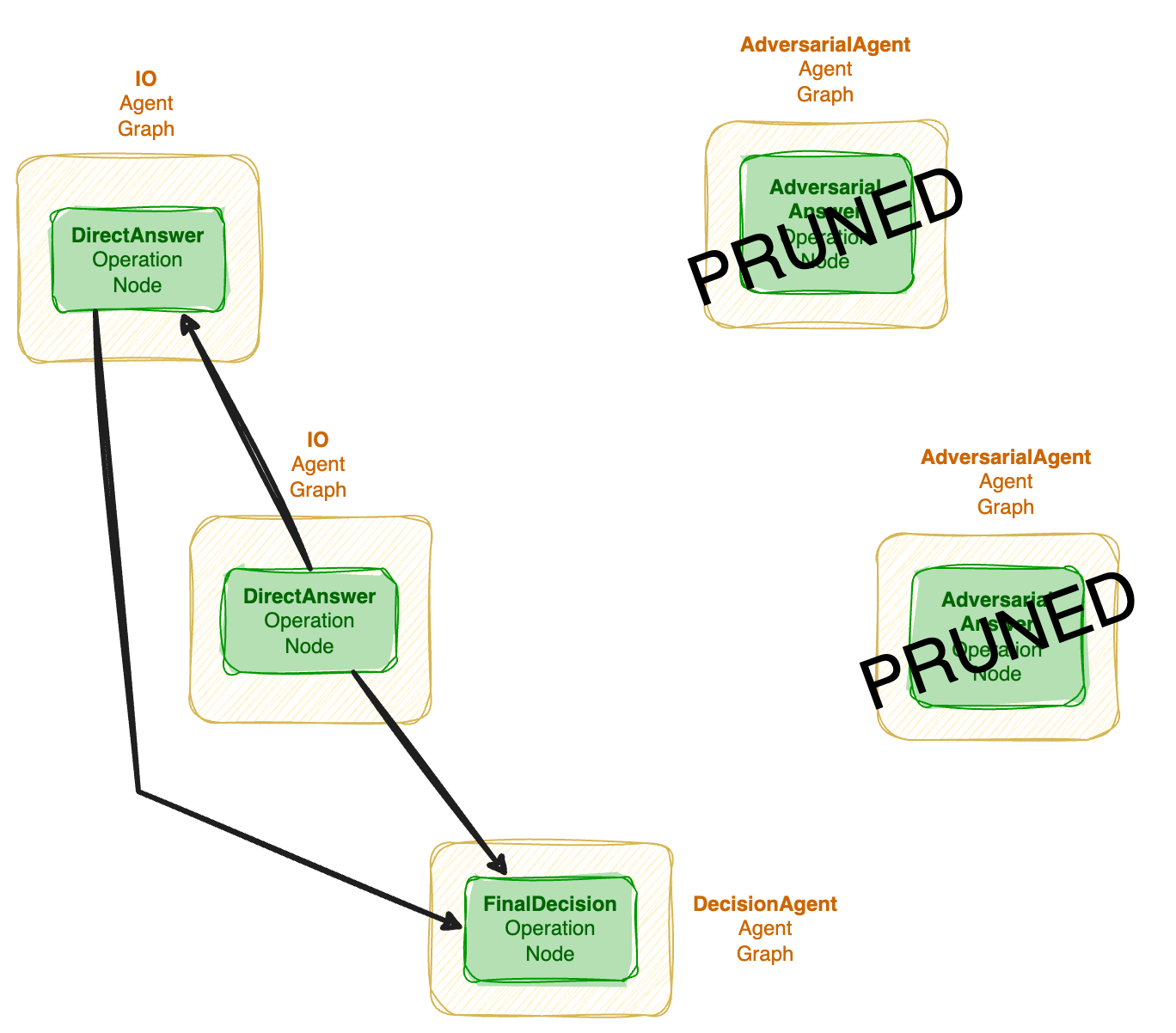}
    \caption{An optimized swarm with realized edges.}
    \label{fig:after_opt}
\end{subfigure}

\caption{A 2T2A swarm before (a) and after (b) optimization}
\label{fig:adv_case}
\end{figure}

An example of a swarm with 2 truthful and 2 adversarial examples is shown in \Cref{fig:adv_case}.
\Cref{fig:before_opt} shows all potential edges before optimization.
\Cref{fig:after_opt} shows only the edges that were connected after optimization was complete.
Note that the disconnected agents and edges are pruned.

\subsubsection{Swarm Roles}
\label{apendix:collab_case}
The list of roles randomly assigned to the IO agents is displayed in \Cref{tab:specialist_roles}.

\begin{table*}[h!]
    \centering
    \caption{Roles that can be assigned to IO agents.}
    \label{tab:specialist_roles}
    
    \renewcommand\tabcolsep{10pt}
    \renewcommand\arraystretch{1.3}
    \footnotesize 
    \begin{tabular}{cccccc}
    \toprule
    Botanist & Data Scientist & Social Worker & Journalist & Pilot \\
    Anthropologist & Fitness Coach & Politician & Artist & Marine Biologist \\
    Ethicist & Entrepreneur & Linguist & Archaeologist & Nurse \\
    Graphic Designer & Philanthropist & Meteorologist & Sommelier & Cybersecurity Expert \\
    \bottomrule
    \end{tabular}

\end{table*}

\subsection{Mini Crosswords}
\subsubsection{Agents Setting}
In our Mini Crosswords experiments, each node returns one or two solutions—either updated or unchanged—for each received solution.
The solutions produced by a node are conditionally independent of each other, given the input solutions of the node.
The output node of a composite graph forwards all received solutions without alteration.
To ensure integration within the system, we mandate the existence of edges from any agent's output node directly to the composite graph's output node.

Our TOT agent uses a tree-search strategy across a perfect binary tree with a depth of eight.
Instead of constructing a graph of $2^9 - 1$ nodes to represent this tree, the search is carried out through a chain of eight branching nodes.
Each branching node is designed to generate two solutions from every input solution that it processes, effectively embodying the TOT strategy.

\subsubsection{Hyper-parameters \& Prompts}
The candidate word generation prompt and the pruning prompt are adapted from the original TOT work~\citep{yao2023tree} and detailed in Table \ref{tab:cw_prompts}.
A clue is defined as a partial filling of the crossword, accompanied by its intended word description and specific position on the board.

\begin{table*}[h!]
  \centering
  \caption{\textbf{Prompts for the Mini Crosswords Experiments.}}
  \label{tab:cw_prompts}
  
  \renewcommand\tabcolsep{10pt}
  \renewcommand\arraystretch{1.2}
  \footnotesize 
  \begin{adjustbox}{scale=0.9}

  \begin{tabular}{|p{0.15\linewidth}|p{0.85\linewidth}|}
    \Xhline{1.2pt}
    \rowcolor{champagne} 
    \textbf{Prompt purpose} & \textbf{Prompt} \\
    \Xhline{1.2pt}
        Candidate words generation prompt &Let's play a 5 x 5 mini crossword, where each word should have exactly 5 letters.

\{current board status\}

Unfilled:

\{Unfilled clues\}

Filled:

\{filled clues\}

Changed:

\{Changed clues\}

Suggestions:

\{suggestions generated by previous Reflection nodes\}

Given the current status, list all possible answers for unfilled or changed words, and your confidence levels (certain/high/medium/low), using the format "h1. apple (medium)". Use "certain" cautiously and only when you are 100\% sure this is the correct word. You can list more then one possible answer for each word.

   \\  \Xhline{1.2pt}
   Pruning prompt
    &  Evaluate if there exists a five letter word of some meaning that fit some letter constraints (sure/maybe/impossible).

Incorrect; to injure: w \_ o \_ g

The letter constraint is: 5 letters, letter 1 is w, letter 3 is o, letter 5 is g.

Some possible words that mean "Incorrect; to injure":

wrong (w r o n g): 5 letters, letter 1 is w, letter 3 is o, letter 5 is g. fit!

sure

A person with an all-consuming enthusiasm, such as for computers or anime: \_ \_ \_ \_ u

The letter constraint is: 5 letters, letter 5 is u.

Some possible words that mean "A person with an all-consuming enthusiasm, such as for computers or anime":

geek (g e e k): 4 letters, not 5

otaku (o t a k u): 5 letters, letter 5 is u

sure

Dewy; roscid: r \_ \_ \_ l

The letter constraint is: 5 letters, letter 1 is r, letter 5 is l.

Some possible words that mean "Dewy; roscid":

moist (m o i s t): 5 letters, letter 1 is m, not r

humid (h u m i d): 5 letters, letter 1 is h, not r

I cannot think of any words now. Only 2 letters are constrained, it is still likely

maybe

A woodland: \_ l \_ d e
The letter constraint is: 5 letters, letter 2 is l, letter 4 is d, letter 5 is e.

Some possible words that mean "A woodland":

forest (f o r e s t): 6 letters, not 5
woods (w o o d s): 5 letters, letter 2 is o, not l

grove (g r o v e): 5 letters, letter 2 is r, not l
I cannot think of any words now. 3 letters are constrained, and \_ l \_ d e seems a common pattern

maybe

An inn: \_ d \_ w f

The letter constraint is: 5 letters, letter 2 is d, letter 4 is w, letter 5 is f.

Some possible words that mean "An inn":
hotel (h o t e l): 5 letters, letter 2 is o, not d

lodge (l o d g e): 5 letters, letter 2 is o, not d

I cannot think of any words now. 3 letters are constrained, and it is extremely unlikely to have a word with pattern \_ d \_ w f to mean "An inn"

impossible

Chance; a parasitic worm; a fish: w r a k \_

The letter constraint is: 5 letters, letter 1 is w, letter 2 is r, letter 3 is a, letter 4 is k.

Some possible words that mean "Chance; a parasitic worm; a fish":

fluke (f l u k e): 5 letters, letter 1 is f, not w

I cannot think of any words now. 4 letters are constrained, and it is extremely unlikely to have a word with pattern w r a k \_ to mean "Chance; a parasitic worm; a fish"

impossible

\{clue\}
 \\
    \Xhline{1.2pt}
    Suggestion Prompt &You are playing a 5 x 5 mini crossword, where each word should have exactly 5 letters.
Given the current status:
\{current board status \}

The target words are classified as Impossible Words, Correct Words, and Incorrect Words.

\texttt{-{}-{}-}

Impossible Words:

\{impossible clues\} 

Correct Words:

\{correct clues \} 

Incorrect Words:

\{incorrect clues \} 

Respond at most five sentences, one sentence per line.
Do not include the phrase "next time" in your response.
       \\  \Xhline{1.2pt}
  \end{tabular}
\end{adjustbox}
\end{table*}

\subsection{HumanEval}

\subsubsection{The Node Optimization Method}
\label{app:humaneval_opt}
For node optimization, we update each node after every four new problem executions. When addressing a new problem $q$ with graph $G$, executing $G(q)$ produces a program, denoted by $s$, whose effectiveness is assessed against test examples associated with $q$. 
The input-output pairs of the nodes generated during the evaluation of $G(q)$ are classified as positive if $s$ passes the tests and as negative otherwise. 
We limit each node to include a maximum of four demonstration examples. 
Let $n$ be a node in the graph associated with a computational routine $f_n^p$, which returns Python programs, parameterized by demonstration examples $p$.
During an optimization step of $n$, that is, an application of $I$ as described in Section \ref{sec:node_opt}, we assess whether to retain existing demonstration examples ($p_n^1$) or to augment them with positive examples from the four most recent problems, subsequently randomly selecting up to four unique examples from this pool (denoted as $p_n^2$).
More specifically, let $Z$ be the set of the last ten inputs of node $n$ received when solving the first-seen problems.
We select $p_n^i$ to update the demonstration examples of $n$, where $i = \argmax_{i \in \{1, 2\}}\sum_{z\in Z}\mathbbm{1}_z(f_n^{p_n^i}(z, q_z))$, $\mathbbm{1}_z$ determines whether a program passes the unit tests stated in $z$, and $q_z$ is the original graph input associated with $z$.

The utility measure for Mini Crosswords experiments is defined as the best state word accuracy, as detailed in Section \ref{sec:crosswords}.
To reduce the variance in gradient estimation with the REINFORCE algorithm, we adjust the utility by subtracting a constant of $0.4$.
For example, a perfectly completed solution results in a utility of $0.6$, while an empty solution yields a utility of $-0.4$.

\subsubsection{Hyper-parameters \& Prompts}
Table \ref{tab:humaneval_prompts} shows the prompts used in our experiments following the principle of ReAct~\citep{yao2022react}.  

\begin{table*}[h!]
  \centering
  \caption{\textbf{Prompts for the Node Optimization experiments on HumanEval.}}
  \label{tab:humaneval_prompts}
  
  \renewcommand\tabcolsep{10pt}
  \renewcommand\arraystretch{1.3}
  \footnotesize 
  \begin{tabular}{|p{0.2\linewidth}|p{0.7\linewidth}|} 
    \Xhline{1.2pt}
    \rowcolor{champagne} 
    \textbf{Prompt purpose} & \textbf{Prompt} \\
    \Xhline{1.2pt}
        System prompt & You are an AI that only responds with only Python code. \\
    \Xhline{1.2pt}
    CodeWriting 
    &  You will be given a function signature and its docstring by the user. Write your full implementation (restate the function signature). Use a Python code block to write your response. For example:
```python
print(`Hello world!')
'''

\{Demonstrations\}

\{problem statement\}
 \\
    \Xhline{1.2pt}
    CodeWriting (ReAct) 
    & You will be given a function signature and its docstring by the user. Write your full implementation (restate the function signature). Use a Python code block to write your response. For example:
```python
print(`Hello world!')
'''

\{Demonstrations\}

Here is an unsuccessful attempt to solve the following question:

Question:

\{problem statement\}

Attempted Solution:

\{previously generated program\}

Feedback:

\{internal unit test results\}

Rewrite the code based on the feedback and the following question:

\{problem statement\} 
 \\
    \Xhline{1.2pt}
  \end{tabular}
\end{table*}

\subsection{GAIA}
\subsubsection{Agent Setting}
We design different agents and swarms. Representative agents and swarms are visualized in \Cref{fig:all_swarm_examples}.
\subsubsection{Hyper-parameters \& Prompts}
We use GPT-4-Turbo for the experiments and design different node operations to solve the GAIA tasks.
\Cref{tab:gaia_prompts1} and \Cref{tab:gaia_prompts2} show the prompts used in our experiments.

\newpage
\begin{table*}[t!]
  \centering
  \caption{\textbf{Prompts for the Task-Solving experiments on GAIA (1).}}
  \label{tab:gaia_prompts1}
  
  \renewcommand\tabcolsep{10pt}
  \renewcommand\arraystretch{1.3}
  \footnotesize 
  \begin{tabular}{|p{0.2\linewidth}|p{0.7\linewidth}|} 
    \Xhline{1.2pt}
    \rowcolor{champagne} 
    \textbf{Prompt purpose} & \textbf{Prompt} \\
    \Xhline{1.2pt}
        System prompt & You are a general AI assistant. I will ask you a question. Report your thoughts, and finish your answer with the following template: FINAL ANSWER: [YOUR FINAL ANSWER]. 
YOUR FINAL ANSWER should be a number OR as few words as possible OR a comma separated list of numbers and/or strings. 
If you are asked for a number, don't use comma to write your number neither use units such as \$ or percent sign unless specified otherwise. 
If you are asked for a string, don't use articles, neither abbreviations (e.g. for cities), and write the digits in plain text unless specified otherwise. 
If you are asked for a comma separated list, apply the above rules depending of whether the element to be put in the list is a number or a string.  \\
    \Xhline{1.2pt}
    DirectAnswer 
    &  \{question\}
   \\ \Xhline{1.2pt} 
    GenerateQuery 
    &  \# Information Gathering for Question Resolution
    
    Evaluate if additional information is needed to answer the question. 

If a web search or file analysis is necessary, outline specific clues or details to be searched for.

\#\# Target Question:

{question}

\#\#  Clues for Investigation:

Identify critical clues and concepts within the question that are essential for finding the answer.
    \\ \Xhline{1.2pt}
    WebSearch 
    &              \# Web Search Task
    
            \#\# Original Question: 
            
            ---
            
            \{question\}
            
            ---
            
            \#\# Targeted Search Objective:
            
            ---
            
            {query}
            
            ---
            
            \#\# Simplified Search Instructions:
            
            Generate three specific search queries directly related to the original question. Each query should focus on key terms from the question. Format the output as a comma-separated list.
            For example, if the question is 'Who will be the next US president?', your queries could be: 'US presidential candidates, current US president, next US president'.
            Remember to format the queries as 'query1, query2, query3'.
    \\ \Xhline{1.2pt}
    DistillWebSearch 
    &   \#\#  Required Information for Summary:

---

\{query\}

---

\#\#  Analyzed Search Results:

---

\{results\}

---

\#\# Instructions for Summarization:

1. Review the provided search results and identify the most relevant information related to the question and query.

2. Extract and highlight the key findings, facts, or data points from these results.

3. Organize the summarized information in a coherent and logical manner.

4. Ensure the summary is concise and directly addresses the query, avoiding extraneous details.

5. If the information from web search is useless, directly answer: \"No useful information from WebSearch\".         
   \\ \Xhline{1.2pt} 
    FileAnalyse 
    &  \# File Analysis Task

\#\#  Information Extraction Objective:

---

\{query\}

---

\#\#  File Under Analysis

---

\{file\}

---

\#\# Instructions:

1. Identify the key sections in the file relevant to the query.

2. Extract and summarize the necessary information from these sections.

3. Ensure the response is focused and directly addresses the query.

Example: 'Identify the main theme in the text.'" 
\\
    \Xhline{1.2pt}
  \end{tabular}
\end{table*}

\newpage
\begin{table*}[t!]
  \centering
  \caption{\textbf{Prompts for the Task-Solving experiments on GAIA (2).}}
  \label{tab:gaia_prompts2}
  
  \renewcommand\tabcolsep{10pt}
  \renewcommand\arraystretch{1.3}
  \footnotesize 
  \begin{tabular}{|p{0.2\linewidth}|p{0.7\linewidth}|} 
    \Xhline{1.2pt}
    \rowcolor{champagne} 
    \textbf{Prompt purpose} & \textbf{Prompt} \\
 \Xhline{1.2pt}
    CombinedAnswer 
    &  
Reference information for FileAnalysis:

---

\{file\_analysis\}

---

Reference information for Websearch:

---
\{web\_search\}
---

Provide a specific answer. For questions with known answers, ensure to provide accurate and factual responses. Avoid vague responses or statements like 'unable to...' that don't contribute to a definitive answer. For example: if a question asks 'who will be the president of America', and the answer is currently unknown, you could suggest possibilities like 'Donald Trump', or 'Biden'. However, if the answer is known, provide the correct information." 
   \\  \Xhline{1.2pt}
   FinalDecision
   
   (Self-Consistency)
    &  \# Self-Consistency Evaluation Task

\#\#  Question for Review:

---

\{question\}

---

\#\#  Reviewable Answers:

---

\{formatted\_answers\}

---

\#\#  Instructions for Selection:

1. Read each answer and assess how it addresses the question.

2. Compare the answers for their adherence to the given question's criteria and logical coherence.

3. Identify the answer that best aligns with the question's requirements and is the most logically consistent.

4. Ignore the candidate answers if they do not give a direct answer, for example, using 'unable to ...', 'as an AI ...'.

5. Copy the most suitable answer as it is, without modification, to maintain its original form.

6. Adhere to the constraints: \{constraint\}.

Note: If no answer fully meets the criteria, choose and copy the one that is closest to the requirements. 
   \\  \Xhline{1.2pt}
   FinalDecision 
   
   (Choose ``Best'')
    &  \#\#  Question:

---

\{question\}

---

\#\#  Candidate Answers for Evaluation:

---

\{formatted\_answers\}

---

\#\#  Evaluation Instructions:

1. Examine the question closely to understand its requirements.

2. Read each candidate answer thoroughly and assess its relevance and accuracy about the question.

3. Choose the answer that most accurately and completely addresses the question.

4. Ignore the candidate answers if they do not give a direct answer, for example, using 'unable to ...', 'as an AI ...'.

"5. Copy the chosen answer exactly as it is presented, maintaining its original format.

6. Adhere to the constraints: 

\{constraint\}.

Note: If none of the answers fully meet the question's criteria, select the one closest to fulfilling them. 
 \\
    \Xhline{1.2pt}
  \end{tabular}
\end{table*}

\section{Resource Requirements}
Table \ref{tab:cost} presents the cost, token consumption, and time requirements of representative experiments. 

\begin{table*}[!htbp]
\centering
\caption{\textbf{Cost, token consumption, and time requirements.} The following experiments are performed with gpt-3.5-turbo-1106 if marked with GPT-3.5T, or gpt-4-1106-preview otherwise.}
  \label{tab:cost}

  \renewcommand\arraystretch{1.2}
  \footnotesize
  \resizebox{\textwidth}{!}{
  \begin{tabular}{l|cccc}
    \Xhline{1.2pt}
    \rowcolor{champagne}
    \textbf{Experiment} & \textbf{Cost (USD)} & \textbf{\# Prompt Tokens} & \textbf{\# Completion Tokens} & \textbf{Time (h)} 
    \\
    \Xhline{1.2pt}
    
    TOT - Mini Crosswords & 65.61 & 1,515,826 & 2,013,511 & 8.5 
    \\
    \rowcolor{gray!10} GPTSwarm - Mini Crosswords Edge-Opt (GPT-3.5T) & 77.42 & 50,394,028 & 13,511,265 & 2.82 
    \\
    GPTSwarm - Mini Crosswords Edge-Opt-Eval (GPT-3.5T) & 9.89 & 6,448,660 & 1,718,613 & 0.73 
    \\
    \rowcolor{gray!10} GPTSwarm - Mini Crosswords Edge-Opt-Eval & 377.54 & 13,137,160 & 8,205,522 & 5.56 
    \\
    GPTSwarm - Mini Crosswords Node-Opt (GPT-3.5T) & 11.22 & 7,468,797 & 1,876,246 & 0.83 
    \\
    \rowcolor{gray!10} GPTSwarm - Mini Crosswords Node-Opt-Eval (GPT-3.5T) & 28.18 & 22,791,158 & 2,693,575 & 0.91 
    \\
    GPTSwarm - HumanEval w/o Opt & 1.61 & 59,646 & 33,951 & 0.68 
    \\
    \rowcolor{gray!10} GPTSwarm - HumanEval w/ Opt & 28.46 & 2,298,140 & 182,594 & 1.49 
    \\
    GPTSwarm - GAIA (Level 1) - Agent(TOT) & 2.21 & 123,801 & 32,599 & 1.05 
    \\
    \rowcolor{gray!10} LLM-Debate - MMLU (3A3T) & 32.8 & 1,689,960 & 530,005 & 8.36 
    \\
    DyLAN - MMLU optimization (3A3T) & 105.93 & 5,671,276 & 1,640,566 & 25.4 
    \\
    \rowcolor{gray!10} DyLAN - MMLU inference (3A3T) & 14.99 & 628,009 & 290,472 & 4.75 
    \\
    GPTSwarm - MMLU optimization (3A3T) & 5.32 & 361,812 & 56,770 & 0.9 
    \\
    \rowcolor{gray!10} GPTSwarm - MMLU inference (3A3T) & 1.82 & 113,233 & 22,923 & 0.31 
    \\
    \Xhline{1.2pt}
  \end{tabular}}
  
\end{table*}

\section{Limitation and Future Work}

In this paper, we focus on optimizing communication between agents, laying the groundwork for more extensive graph optimization in the future.
For example, while current methods optimize edge connections between agents, the internal node topology of each agent is also crucial.
Dynamically changing the topology may enhance task planning.
Additionally, scaling up the agent framework is essential.
When the number of agents exceeds 100, maintaining communication efficiency and system robustness becomes a significant challenge.

\end{document}